\documentclass[manuscript,screen]{acmart}
\usepackage{multirow}

\usepackage{algorithm}
\usepackage[noend]{algorithmic}
\makeatletter
\newcommand{\ALOOP}[1]{\ALC@it\algorithmicloop\ #1%
  \begin{ALC@loop}}
\newcommand{\ENDALOOP}{\end{ALC@loop}\ALC@it\algorithmicendloop}

\makeatother

\newcommand{\bg}[1]{\boldsymbol{#1}} 
\newcommand{\bm}[1]{\mathbf{#1}} 

\newcommand\raiseT[2]{%
\setbox0\hbox{$#1{#2}$}\raise\dp0\box0}

\AtBeginDocument{%
  \providecommand\BibTeX{{%
    \normalfont B\kern-0.5em{\scshape i\kern-0.25em b}\kern-0.8em\TeX}}}

\setcopyright{acmcopyright}
\copyrightyear{2021}
\acmYear{2021}
\acmDOI{10.1145/1122445.1122456}

\begin{document}

\title{A Federated Learning Approach to Anomaly Detection in Smart Buildings}

\author{Raed Abdel Sater}
\email{raed.abdelsater@mail.concordia.ca}
\author{A. Ben Hamza}
\authornotemark[1]
\email{hamza@ciise.concordia.ca}
\affiliation{%
  \institution{Concordia Institute for Information Systems Engineering}
  \streetaddress{Concordia University, 1455 de Maisonneuve Blvd. West, EV7.640}
  \city{Montreal}
  \state{QC, Canada}
  \postcode{H3G 1M8}
}

%

\begin{abstract}
Internet of Things (IoT) sensors in smart buildings are becoming increasingly ubiquitous, making buildings more livable, energy efficient, and sustainable. These devices sense the environment and generate multivariate temporal data of paramount importance for detecting anomalies and improving the prediction of energy usage in smart buildings. However, detecting these anomalies in centralized systems is often plagued by a huge delay in response time. To overcome this issue, we formulate the anomaly detection problem in a federated learning setting by leveraging the multi-task learning paradigm, which aims at solving multiple tasks simultaneously while taking advantage of the similarities and differences across tasks. We propose a novel privacy-by-design federated learning model using a stacked long short-time memory (LSTM) model, and we demonstrate that it is more than twice as fast during training convergence compared to the centralized LSTM. The effectiveness of our federated learning approach is demonstrated on three real-world datasets generated by the IoT production system at General Electric Current smart building, achieving state-of-the-art performance compared to baseline methods in both classification and regression tasks. Our experimental results demonstrate the effectiveness of the proposed framework in reducing the overall training cost without compromising the prediction performance.
\end{abstract}


\maketitle
\section{Introduction}
Smart buildings are making bold use of data collected by IoT sensors to assist in a wide range of positive outcomes, including cost reduction, improved safety and maintenance, and prevention of building equipment downtime. In simple terms, IoT refers to a network of sensors and other devices that are capable of sending and receiving data~\cite{Perera:15,Mosenia:17}, allowing diverse network components to cooperate and make their resources available to execute a common task. IoT helps create interoperable networks in smart buildings by connecting various types of sensors and other devices from which actionable insights can be extracted through collection and analysis of massive amounts of real-time data.

IoT-enabled smart buildings are changing the way we live, leveraging the power of intelligent devices to remotely monitor and control key equipment in the premises while ensuring greater energy and operational efficiency~\cite{ma2016multi,Khamesi:20}. In such an intelligent environment, a key objective is to provide support tools to help building managers and users make cost-effective decisions when utilizing, for example, electrical energy. It is estimated that more than half the global electrical energy is consumed by commercial buildings, and around 45\% of that energy is generated by Heating, Ventilation and Air Conditioning (HVAC) systems~\cite{d2018human}. In order to intelligently improve performance and create smarter buildings, there is a pressing need for developing efficient machine learning models that effectively learn the history patterns of IoT sensors in an effort to increase energy efficiency and help cut costs. Such models not only help facility managers make strategic decisions through data analysis and actionable insights to ensure buildings are working smarter and running at maximum efficiency, but can also help buildings self-diagnose and optimize. In particular, an anomaly detection model can be used to deliver insights on the present and future performance of critical assets in smart buildings.

Anomaly detection is the process of identifying anomalous observations, which do not conform to the expected pattern of other observations in a dataset. Detecting anomalies has become a central research question in IoT applications, particularly from IoT time-series data~\cite{HKim:20,Pajouh:19,cook2019anomaly}. For instance, a lighting energy consumption pattern might vary in an office building. Smart lighting control systems use occupancy sensors in correlation with light sensors to dim light based on changing occupancy and daylight levels, with the benefit of conserving energy in the building~\cite{violatto2019classification}. Moreover, an observation of high energy consumption levels due to heating in a bank might be anomalous after working hours, but not during the day. Occupancy sensors can be used to measure occupancy and space utilization, as well as in other systems such as HVAC to detect such anomalies~\cite{beltran2014optimal}. Multiple types of sensors are of great importance to building operations and can be used for a number of applications in an IoT environment, including energy consumption monitoring and control, and security of critical systems. Most modern structures are equipped with a building automation system (BAS), which enables facility managers to automate and oversee the energy efficiency of a building by controlling various components within a building's structure, such as HVAC.

Early works on detection and prediction tasks in smart buildings have focused on centralized approaches using IoT sensor data. Id\'e \textit{et al.}~\cite{ide2007computing} used correlation between multiple sensors to form neighborhood graphs for computing correlation anomaly scores, but they treat a multi-sensor system as a collection of centralized sensors. Bellido-Outeirino~\textit{et al.}~\cite{bellido2012building} presented a building lighting automation system by integrating digital addressable lighting interface devices in wireless sensor networks using a centralized system, in which appliances are managed by a wireless sensor network that focuses on lighting automation, while considering maintenance and energy consumption costs. Yu~\textit{et al.}~\cite{yu2017distributed} developed a centralized real-time HVAC system to construct and stabilize virtual queues associated with indoor temperatures by minimizing the long-term total cost associated with the HVAC system in the smart grid, which is considered a key IoT application that involves the incorporation of a secure, two-way information and communication flow, along with a two-way power flow. Li \textit{et al.}~\cite{li2018leak} proposed a centralized water leak detection method using a classifier based on artificial neural networks using acoustic emission sensor data in an effort to detect water leaks in municipal pipeline systems. Chandra \textit{et al.}~\cite{chandra2018bayesian} presented a Bayesian approach to multi-task learning for dynamic time series prediction via cascaded neural networks by decomposing a single task learning problem into multi-task learning through subtasks that have inter-dependencies defined by the size of the window used for embedding. In order to detect spatial, temporal, and spatio-temporal anomalies in real-time sensor measurement data streams, Chen \textit{et al.}~\cite{chen2017adf} proposed an anomaly detection framework for real-world environmental sensing systems in a bid to identify outliers in raw measurement data.

In centralized systems, an abnormal sensor behavior is detected by a central model on a server. Typically, all the training data are collected from the sensors through gateways and saved on the cloud. Centralized systems are, however, prone to failures, vulnerable to cyber invasions and infections, and often require longer access times for training data coming from remote devices, leading to potential data exposure during the transmission from clients to server, and hence raising concerns about clients' privacy~\cite{goodhue1991security}. Federated learning has recently emerged as a powerful alternative to centralized systems, as it enables the collaborative training of machine learning models from decentralized datasets in users' devices such as mobile phones, wearable devices and smart sensors without uploading their privacy-sensitive data to a central server or service provider~\cite{KonecnyMYRSB16,mcmahan2016communication,Mehryar:19,Niyato:19,TianLi:19,YChen:19,KBonawitz:19,PLiang:20,Hongyi:20}, while reducing communication cost~\cite{bonawitz2017practical}. For example, in a federated learning system for smart buildings, each sensor performs model training using its own data and sends local updates to the central server for aggregation in order to update the global model, which is then redistributed to the sensors. The training process of a federated learning model is usually carried out using the federated averaging algorithm~\cite{mcmahan2016communication}, and is repeated iteratively until model convergence or the maximum number of training rounds is reached. Moreover, secure aggregation can be used to combine the outputs of local models on sensors for updating a global model by leveraging secure multiparty computation~\cite{brendan2017learning}.

While federated learning has some privacy-enhancing advantages as compared to sharing private data with a central server, recent studies have shown that an honest-but-curious server can analyze the local model parameter updates to perform gradient leakage attacks~\cite{zhu2019dlg,Jonas2020Geiping}, which usually occur on comprised sensors, and hence the server may gain access to some private training data. To mitigate this issue of data leakage, federated learning can be used in conjunction with secure multi-party computation, homomorphic encryption, or differential privacy. With secure multi-party computation, for instance, the local model updates are securely combined into a single aggregate update in order to ensure that the data communicated through federated learning to the centralized server stays private~\cite{bonawitz2017practical}. Another effective strategy to prevent the gradients from leaking private training data is to set those with small magnitudes to zero~\cite{zhu2019dlg}.

In this paper, we introduce a federated stacked long short-time memory model using federated learning on time series data generated by IoT sensors for classification and regression tasks such as lighting fault detection and energy usage prediction. In addition to learning long-term dependencies between time steps of sequence data, the proposed model learns individual feature correlations within each sensor and also shared feature representations across distributed datasets to improve model convergence for faster learning. We perform the training process, which involves input data from multiple heterogeneous data sources across numerous sensors, in a collaborative fashion on IoT sensors in lieu of the server. This strategy not only lowers the cost and anticipates failures, but also helps reduce computational demands and inherits the privacy-enhancing capabilities of federated learning. The proposed framework provides: (i) a reduced network traffic by sharing the weights solely with the federated server, (ii) a shorter convergence time thanks to the collaborative learning, and (iii) a federated methodology for training various types of sensors such as occupancy sensors. Moreover, we design a federated gated recurrent unit model and also a federated logistic regression to learn a common representation among multiple sensors. Our main contributions can be summarized as follows:
\begin{itemize}
\item We leverage the multi-task learning paradigm to formulate the anomaly detection problem in smart buildings.
\item We propose a novel privacy-by-design federated stacked long short-time memory model for anomaly detection in smart buildings using IoT sensor data.
\item We demonstrate that our proposed federated stacked LSTM model converges 2x faster than the centralized LSTM during the training phase and significantly reduces the communication cost.
\item We present experimental results to demonstrate the superior performance of our model in comparison with baseline methods on sensors event log and energy usage datasets.
\end{itemize}

\medskip\noindent
The rest of this paper is organized as follows. In Section 2, we provide a brief overview of centralized and federated learning approaches for anomaly detection in a smart building setting using IoT sensor data. In Section 3, we present our federated learning system as well as our problem formulation and propose a novel federated learning architecture for anomaly detection in smart buildings. We discuss in detail the main components and algorithmic steps of the proposed framework. In Section 4, we present experimental results to demonstrate the superior performance of our approach in comparison with baseline methods. Finally, we conclude in Section 5 and point out future work directions.

\section{Related Work}
The growing number of smart city centers has accelerated the proliferation of IoT sensors to observe and/or interact with their internal and external environments. In recent years, the advent of deep learning has sparked interest in the adoption of deep neural networks for learning latent representations of IoT sensor data. Du \emph{et al.}~\cite{MinDu:17} introduced a deep neural network using LSTM to model a system log as a natural language sequence by learning log patterns from normal execution in order detect anomalies when log patterns deviate from the model trained from log data under normal execution. Zhu \emph{et al.}~\cite{zhu2019mobile} proposed a multi-task anomaly detection framework for control area network messages using LSTM on in-vehicle network data. Zhang \emph{et al.}~\cite{zhang2018lstm} also used an LSTM model to predict power station working conditions from data generated by industrial IoT sensors of a main pump in a power station. However, all the aforementioned approaches are centralized, as data generated by IoT devices are sent directly to the server, raising privacy concerns and resulting in network overload. Moreover, the training process of centralized approaches is bandwidth intensive and comes with significant privacy implications.

More recently, federated learning has emerged as a viable, compelling alternative to the centralized approach. Rather than aggregating increasingly large amounts and types of data into a central location, federated learning distributes the global model training process such that each participating sensor's data is used in situ to train a local model. Nguyen \emph{et al.}~\cite{nguyen2019diot} proposed an anomaly detection system for detecting compromised IoT devices using federated learning by aggregating anomaly-detection profiles for intrusion detection. Li \emph{et al.}~\cite{li2019abnormal} presented an autoencoder based anomaly detection approach at the server side to detect anomalous local weight updates from the clients in a federated learning system. Yurochkin \emph{et. al.}~\cite{Yurochkin:19} developed a probabilistic federated learning framework with a particular emphasis on training and aggregating neural network models by decoupling the learning of local models from their aggregation into a global federated model. Zhao \emph{et. al.}~\cite{YZhao:19} presented a multi-task federated learning method for computer networks anomaly detection, as well as for traffic recognition and classification tasks. Chen \emph{et. al.}~\cite{YChen:19} introduced a federated multi-task hierarchical attention model for activity recognition and environment monitoring using multiple sensors.

\section{Method}
Motivated by LSTM networks' capability of learning long-term dependencies between time steps of sequence data~\cite{HochSchm97}, we introduce in this section a novel privacy-by-design federated stacked LSTM model for anomaly detection in smart buildings using IoT sensor data. The proposed approach leverages the multi-task learning paradigm.

\begin{figure}[!htb]
\centering
\includegraphics[scale=0.3]{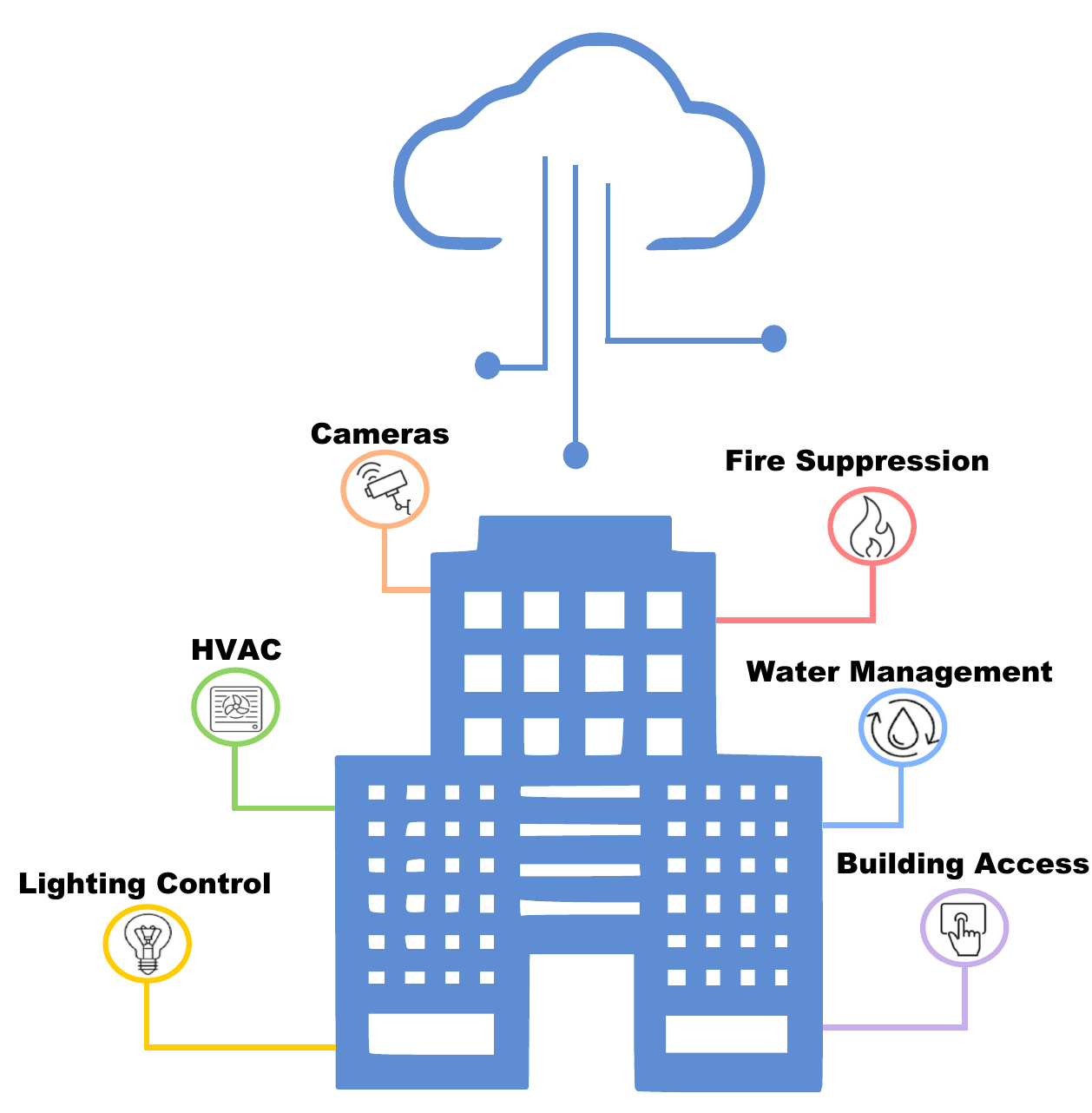}
\caption{IoT-enabled smart building.}
\label{Fig:Smart_Building}
\end{figure}

In a smart building setting with a federated learning system setup, we assume that there are $K$ sensors, each of which has a dataset $\mathcal{D}_k$ that is kept private, where $k=1,\dots,K$. In other words, the data $\mathcal{D}_k$ of the $k$-th sensor are not shared with the server. An IoT-enabled smart building is depicted in Figure~\ref{Fig:Smart_Building}, which shows several types of sensors for a variety of tasks, including lighting control, building access, water management, HVAC, fire suppression, and building monitoring. Unlike the traditional centralized learning that collects and uses all local data $\mathcal{D}=\cup_{k=1}^{K}\mathcal{D}_k$ from all sensors to train a model, the federated learning paradigm only collects and aggregates updated local models from the sensors to generate a global model.

\begin{figure}[!hb]
\centering
\includegraphics[scale=.41]{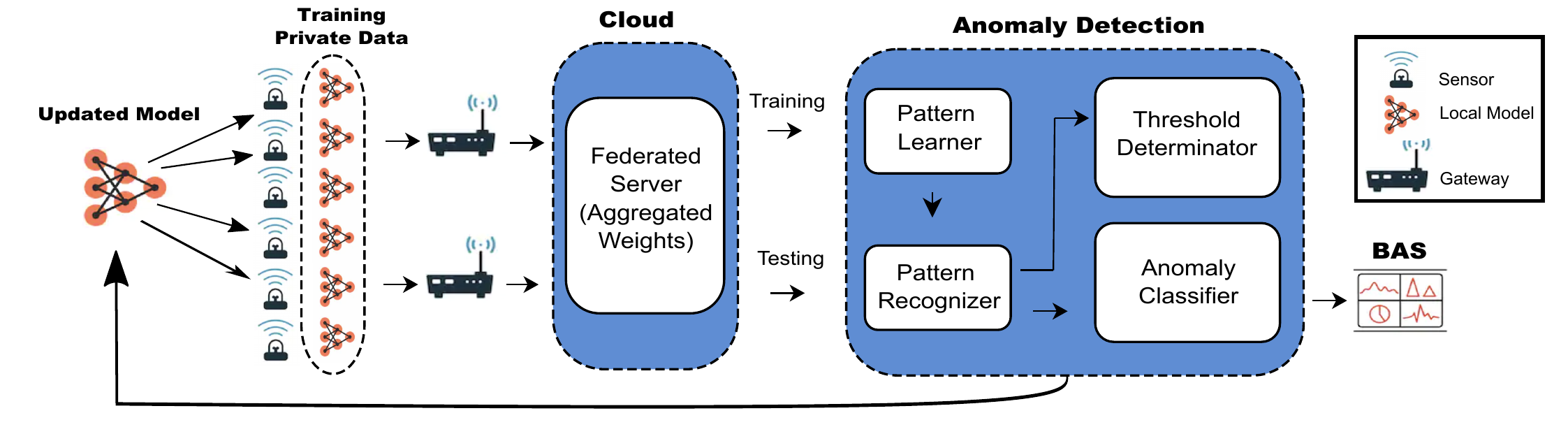}
\caption{Federated learning system for anomaly detection.}
\label{Fig:Federated_system}
\end{figure}

An overall schematic layout of our federated learning system for anomaly detection is illustrated in Figure~\ref{Fig:Federated_system}. The proposed system uses historic sensor data and contextual features to identify temporal anomalies in smart buildings using a multi-task federated recurrent neural network to classify anomalous sensors and predict energy consumption. The main building blocks of this system are local training, cloud aggregation, anomaly detection and global model broadcast.

\medskip\noindent{\textbf{Local training.}}\quad Sensor data represent a time stamped input of energy consumption data recorded at regular time intervals. These sensor data are often noisy and incomplete due largely to faulty devices and/or communication errors~\cite{mitchell2007multi}. To mitigate the negative impact of noisy and incomplete data on the performance of the federated learning system, these data instances are usually discarded or interpolated. Each sensor trains the local model on anomalous instances. The basic assumption is that historic sensor data are predominantly normal. The historical real datasets can then be split into training and test sets with the objective of using the test data to evaluate the capacity of the anomaly detection framework in order to identify normal behavior during the anomaly detection phase. The primary goal of training is to enhance engine recognition of normal input data patterns. At the end of the training phase, the participating sensors send the learned parameters of their local models for aggregation to a central server via Wireless Area Controller gateways that monitor and control data traffic between the sensors and the IoT system in the cloud~\cite{kazemi2011novel}.

\medskip\noindent{\textbf{Cloud aggregation.}}\quad We use PySyft, a Python library for secure and private deep learning~\cite{PySyft:18}, which integrates federated learning with secure multi-party computation and differential privacy in an effort to protect against threats within the data center by ensuring that individual sensors' updates remain encrypted in memory. The secure multi-party computation protocol leverages encryption to make individual sensors' updates uninspectable by a server~\cite{bonawitz2017practical}. At each communication round between the server and participating sensors, the server aggregates the learned parameters of the local models using the Federated Averaging algorithm, followed by updating the global model. This training process is repeated for a certain number of communication rounds until a desirable level of performance is achieved. Then, the updated global model is prepared for integration into the anomaly detection engine.

In PySyft, private data leakage is mitigated with secure aggregation~\cite{bonawitz2017practical} by leveraging secure multiparty
computation to compute sums of local model parameter updates from individual sensors while maintaining privacy guarantees. With secure aggregation, the local model parameter updates are kept encrypted and their sum is only revealed to the server after a sufficient number of communication rounds. Homomorphic encryption and differential privacy can also be used in PySyft. However, only secure aggregation is used in our algorithm. In differential privacy, for instance, each sensor adds a carefully calibrated amount of noise to its local parameter update in an effort to mask its contribution to the learned global model.

\medskip\noindent{\textbf{Anomaly detection.}}\quad The anomaly detection block of the proposed system is composed of four main components: the pattern learner process, pattern recognizer, threshold determinator, and anomaly classifier. The pattern learner receives the global model after the training has been completed and configured for testing. The pattern recognizer tests the global model and evaluates the performance based on the test data to help the threshold determinator find a suitable value for the specific problem. Based on the value of the threshold, the anomaly classifier generates the inference and predicts the anomalous data points that did not pass the threshold determinator and flags them in the anomaly detection system. The updated global model will then be ready for deployment on sensors after this round of training.

\medskip\noindent{\textbf{Global model broadcast.}}\quad Once the model weights are updated using the Federated Averaging algorithm, the anomaly detection engine, located in the IoT system, sends back the updated global model to all sensors that are involved in the first round of training. These sensors receive the updated global model with new weights and then replace their local models' parameters with the global parameters to start a new round of training.

\subsection{Multi-Task Learning}
We leverage the multi-task learning paradigm, which aims at solving multiple tasks simultaneously while taking advantage of the similarities and differences across tasks, to formulate the anomaly detection problem in a federated learning setting. Our notation is summarized in Table~\ref{tab:TableOfNotations}.

\begin{table}[!htb]
\caption{Notation.}
\begin{tabular}{l c p{7cm} }
\toprule
\textbf{Notation} & & \textbf{Description}\\
\midrule
$K$ &  & Number of participating sensors indexed by $k$\\
$\bm{X}^{k}$ &  & Training data for the $k$-th sensor with $N_k$ local examples\\
$\bm{y}^k$ & & Ground-truth label or target output vector\\
$\hat{\bm{y}}^k$ & & Predicted probability or response vector\\
$ \mathcal{L}(\cdot,\cdot)$ & & Loss function \\
$\sigma(\cdot)$ &  & Sigmoid activation function\\
$B$ & &  Local minibatch size \\
$E$ & &  Number of local epochs \\
$w_G$ &  & Global model\\
FC &  & Fully Connected layer\\
BA &  & Balanced Accuracy\\
LR & & Logistic Regression \\
FLR & & Federated Logistic Regression \\
FGRU & & Federated Gated Recurrent Unit \\
LSTM & & Long Short-Term Memory network \\
$\textrm{FLSTM-}\ell$ & & Federated LSTM with $\ell$ layers \\
FSLSTM & & Federated Stacked LSTM (i.e. FLSTM-3)\\
\bottomrule
\end{tabular}
\label{tab:TableOfNotations}
\end{table}

Suppose that we have $K$ learning tasks (i.e. one task per sensor) and denote by $(\bm{X}^{k},\bm{y}^k)$ the training data for the $k$-th task, where $\bm{X}^k =(\bm{x}_{1}^{k},\dots, \bm{x}_{N_k}^{k})^{T}\in \mathbb{R}^{{N_k}\times F}$ is a data matrix consisting of $N_k$ samples collected from multiple sensors of the same category (e.g. occupancy sensors), and $\bm{y}^{k}=(y_{1}^{k},\dots,y_{N_k}^{k})^{T} \in\mathbb{R}^{N_k}$ is a vector of outputs for the $N_k$ samples. The training samples in $\bm{X}^k$ are generated by sensors connected to different gateways in the building. Further, we assume that the samples have the same feature dimension $F$ across all tasks. It is important to note that $y_{i}^k\in\mathbb{R}$ for regression tasks, while $y_{i}^k\in\{0,1\}$ for binary classification tasks with 0 and 1 representing ``normal'' and ``anomalous'' observations, respectively.

The goal of multi-task learning is to learn the weight parameters of a model by minimizing the following objective function across all sensors
\begin{equation}
\mathcal{E}_{\text{avg}} = \frac{1}{K}\sum_{k=1}^{K}\mathcal{L}(\bm{y}^k,\hat{\bm{y}}^k),
\end{equation}
where $\mathcal{L}$ is a loss function and $\hat{\bm{y}}^k$ is a vector of predicted values by the model for the $k$-th task.

For binary classification tasks, we train a model to minimize the cross-entropy loss function given by
\begin{equation}
\mathcal{L}(\bm{y}^k,\hat{\bm{y}}^k) = -\frac{1}{N_{k}}\sum_{i=1}^{N_k}y_{i}^k\log(\hat{y}_{i}^k),
\end{equation}
where $y_{i}^k$ and $\hat{y}_{i}^k$ are the ground-truth label and predicted probability, respectively, for the $k$-th task.

For regression tasks, we train a model to minimize the mean squared error (MSE) given by
\begin{equation}\label{eq:MSE}
\mathcal{L}(\bm{y}^k,\hat{\bm{y}}^k) = \frac{1}{N_{k}}\sum_{i=1}^{N_k}(y_{i}^k - \hat{y}_{i}^k)^2,
\end{equation}
where $y_{i}^k$ and $\hat{y}_{i}^k$ are the target output and predicted value by the model, respectively, for the $k$-th task.

\subsection{LSTM}\label{sec:lstm}
\begin{figure}[!htb]
\centering
\includegraphics[scale=0.65]{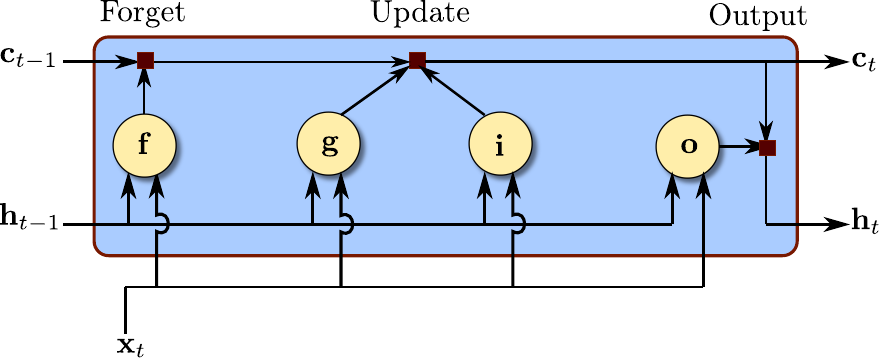}
\caption{LSTM block architecture.}
\label{Fig:LSTM_Architecture}
\end{figure}

LSTM networks are a special type of recurrent neural networks (RNNs), capable of learning long-term dependencies between time steps of sequence data while being resilient to the vanishing gradient problem~\cite{HochSchm97}. The key to an LSTM network is the cell state, which contains information learned from the previous time steps and has the ability to remove or add information using gates~\cite{chen2018meta}. These gates control the flow of information to and from the memory. In addition to the hidden state, the architecture of an LSTM block is composed of a cell state, forget gate, memory cell, input gate and output gate, as illustrated in Figure~\ref{Fig:LSTM_Architecture}. At each time step, the LSTM block takes as input the current input data vector $\bm{x}_t$ and both the hidden state (i.e. short-term memory) $\bm{h}_{t-1}$ and cell state (i.e. long-term memory) $\bm{c}_{t-1}$ from the previous cell. In order to decide which information to be retained or discarded at each time step before passing on the long-term and short-term information to the next cell, the LSTM block uses the forget, input and output gates, which are trainable functions with weights and biases. The forget gate decides which information from the long-term memory to forget, while the input gate can be regarded as a filter that selects what information can be kept and what information to be thrown out. The memory cell $\bm{g}_t$ is created by passing the current input and short-term memory into a $\tanh$ activation function, which is a shifted version of the sigmoid activation function. The new cell state $\bm{c}_{t}$ is obtained by adding two pointwise multiplication terms; the first term involves the input gate and memory cell, while the second one uses the forget gate and the previous cell state. The cell state $\bm{c}_{t}$ stores information about the input data across time steps. Finally, the hidden state $\bm{h}_t$ is obtained via pointwise multiplication of the output gate $\bm{o}_t$ and the new cell state through a $\tanh$ activation function. This hidden state (i.e. new short-term memory) is then passed on to the cell in the next time step.

Formally, given the input $\bm{x}_t$, current cell state $\bm{c}_{t-1}$ and hidden state $\bm{h}_{t-1}$ of the network, the LSTM updates at time step $t$ are given by

\begin{align}
\begin{split}
\bm{f}_t &= \sigma(\bm{W}_{f}\bm{x}_{t} + \bm{R}_{f}\bm{h}_{t-1}+\bm{b}_{f})\\
\bm{g}_t &= \tanh(\bm{W}_{g}\bm{x}_{t} + \bm{R}_{g}\bm{h}_{t-1}+\bm{b}_{g})\\
\bm{i}_t &= \sigma(\bm{W}_{i}\bm{x}_{t} + \bm{R}_{i}\bm{h}_{t-1}+\bm{b}_{i})\\
\bm{c}_t &= \bm{f}_{t}\odot\bm{c}_{t-1}+\bm{i}_{t}\odot\bm{g}_{t}\\
\bm{o}_t &= \sigma(\bm{W}_{o}\bm{x}_{t} + \bm{R}_{o}\bm{h}_{t-1}+\bm{b}_{o})\\
\bm{h}_t &= \bm{o}_{t}\odot\tanh(\bm{c}_{t})
\end{split}
\label{eq:lstm}
\end{align}
where $\bm{f}_t$, $\bm{g}_t$, $\bm{i}_t$, $\bm{o}_t$, $\bm{c}_t$ and $\bm{h}_t$ are the forget gate, memory cell, input gate, output gate, cell state and hidden state, respectively; $\sigma(\cdot)$ denotes the sigmoid activation function; $\odot$ denotes the point-wise product; $\bm{W}_{\bullet}$ and $\bm{R}_{\bullet}$ are the learnable input and recurrent weight matrices; and $\bm{b}_{\bullet}$ are the learnable bias vectors.

In summary, the input gate controls what new information is added to cell state from current input, while the forget gate controls what information to throw away from memory. The output gate controls what information encoded in the cell state is sent to the network as input in the following time step. An LSTM network with multiple LSTM layers is referred to as a stacked or deep LSTM, with the output sequence of one LSTM layer forming the input sequence of the next.

\subsection{Proposed Framework}
\begin{figure}[!htb]
\centering
\includegraphics[scale=0.23]{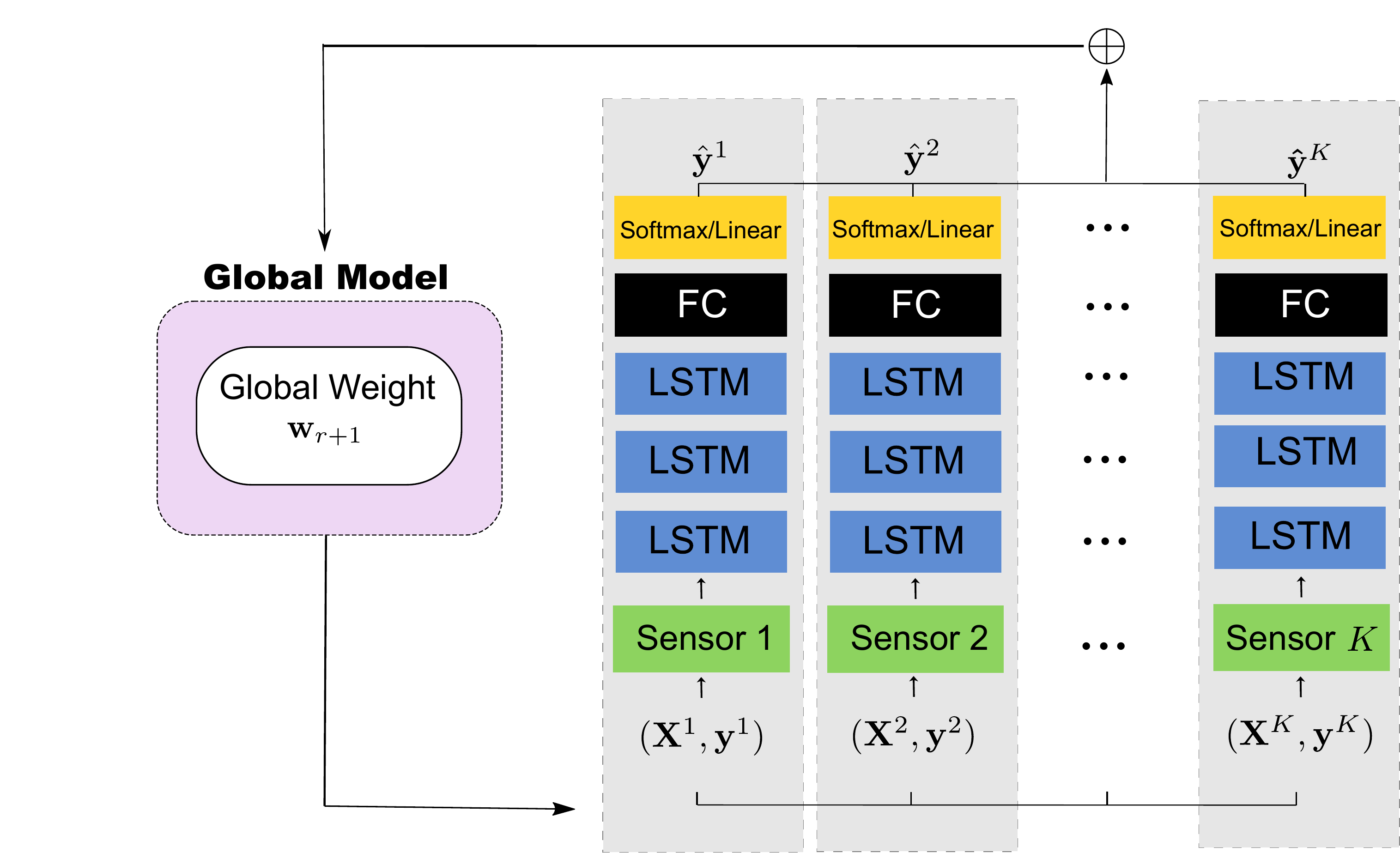}
\caption{Architecture of proposed federated stacked LSTM model. Only the local model parameter updates are sent back to the server.}
\label{Fig:Federated_Stacked_LSTM}
\end{figure}

In order to overcome the centralized training issues such as computational demand, IoT device availability and network bandwidth limitation, we introduce a federated stacked long short-time memory (FSLSTM) model that contains two main components: a local model and a global model. The local model is a stacked LSTM network composed of three LSTM layers, and is applied on input data generated by each sensor to learn a latent feature representation. A fully connected (FC) layer is applied to the hidden representation of the last LSTM layer, followed by a softmax or linear activation function for classification and regression tasks, respectively. The global model, on the other hand, aggregates the weights of all local models after each round. The architecture of the proposed FSLSTM model consists of three LSTM layers, as illustrated in Figure~\ref{Fig:Federated_Stacked_LSTM}.

In a federated learning setting, the training of a stacked LSTM model is distributed across the participating sensors by iteratively aggregating local models into a joint global model. Suppose we have training local datasets $\{(\bm{X}^{k},\bm{y}^k)\}_{k=1}^{K}$ from $K$ sensors, where the total number of samples distributed over these sensors is $N=\sum_{k=1}^{K}N_k$, with $N_k$ denoting the number of samples in the training set $\mathcal{D}_{k}=(\bm{X}^{k},\bm{y}^k)=\{(\bm{x}_{i}^{k},y_{i}^{k})\}_{i=1}^{N_{k}}$ of the $k$-th sensor. We consider a federated learning task, where these $K$ sensors collaboratively train a model parameter vector $\bm{w}$ with the orchestration of a remote server. The goal is to minimize the following global loss function on all the distributed datasets
\begin{equation}
f(\bm{w})=\sum_{k=1}^{K}\frac{N_{k}}{N}f_{k}(\bm{w}),
\end{equation}
where $f_{k}(\bm{w})$ is the local loss function on the collection of data samples $\bg{\xi}_{i}^{k}=(\bm{x}_{i}^{k},y_{i}^{k})$ from the $k$-th sensor
\begin{equation}
f_{k}(\bm{w})=\frac{1}{N_{k}}\sum_{i=1}^{N_{k}}\ell(\bm{w};\bg{\xi}_{i}^{k}),
\end{equation}
and $\ell(\bm{w};\bg{\xi}_{i}^{k})$ is the loss function of the prediction on the data sample $\bg{\xi}_{i}^{k}$ made with model parameter vector $\bm{w}$. It is important to point out that in the case of the IID assumption (i.e. the training examples are distributed over the clients uniformly at random), we have $\mathbb{E}_{\mathcal{D}_{k}}[f_{k}(\bm{w})]=f(\bm{w})$, where the expectation is taken over the set of samples assigned to a fixed sensor $k$.

Each local LSTM model receives a local copy of the global model weights based on a time series window with step $t$.

\medskip\noindent{\textbf{Local model.}}\quad Each participating sensor uses its local data to update parameters of the local model, which is a stacked LSTM network with three layers. As stated earlier, at each time step $t$, the LSTM block takes as input the current input data vector $\bm{x}_t$ and both the hidden state $\bm{h}_{t-1}$ and cell state $\bm{c}_{t-1}$ from the previous cell. Then, the LSTM network learns to predict the hidden and cell states of the next time step as follows:
\begin{equation}
\bm{h}_{t}, \bm{c}_{t} = \text{LSTM}(\bm{h}_{t-1},\bm{x}_{t}, \bm{c}_{t-1}; \bm{w}),
\end{equation}
where LSTM is an operator representing the operations in Eq.~\eqref{eq:lstm}, $\bm{h}_{t}$ is the new hidden state, $\bm{c}_{t}$ is the new cell state, and $\bm{w}$ is the parameter vector of the LSTM model. We initialize $\bm{c_0}$ and $\bm{h_0}$ as random vectors.

The output of the last hidden state is the hidden vector of the last time step, and can be viewed as the representation of the whole sequence. Then, we pass this hidden representation to a fully-connected layer to get the predicted values for the $k$-th task as follows:
\begin{equation}
\hat{\bm{y}}^{k} = \sigma(\bm{W}^{\text{FC}} \bm{h}_{\text{last}}+ \bm{b}^{\text{FC}}),
\end{equation}
where $\sigma$ is the softmax or linear activation function, $\bm{h}_{\text{last}}$ denotes the hidden vector in the last time step of LSTM, $\bm{W}^{\text{FC}}$ and $\bm{b}^{\text{FC}}$ are the learnable weight matrix and bias vector of the FC layer.

\medskip
\noindent{\textbf{Global model.}}\quad At each round $r$, the server randomly chooses a subset $S_r$ of sensors for synchronous aggregation and broadcasts the global model parameter vector $\bm{w}_{r}$ to the participating sensors. For the choice of $S_r$, the random number generator is initialized with the same seed value for all training rounds. Each sensor updates the local model parameters by minimizing the loss function over its local data dataset, starting from the global model parameter vector $\bm{w}_{r}$ shared by the server and using stochastic gradient descent algorithm for $E$ epochs and with a batch size $B$. At the end of the training phase, the parameters of local models are sent to the server for aggregation, which is done synchronously using the federated averaging algorithm in order to obtain the global model parameter vector $\bm{w}_{r+1}$ for the next round.

The update of local and global models is repeated for a certain number of rounds until the global loss function converges. The main algorithmic steps of our proposed federated stacked LSTM network are summarized in Algorithm 1. This algorithm starts by randomly initializing the global model. Then, the server randomly selects a subset of sensors and distributes the current global model to these sensors. Each sensor trains the global model with its local data independently, and then the server collects the parameters of the locally trained models for all selected sensors and aggregates them using the federated averaging algorithm by computing a weighted average of these parameters in order to obtain a shared global model.

At the beginning of the training round of communication, each sensor reads the current parameter vector of the global model from the central server and updates it via stochastic gradient descent, where the stochastic gradient computed using a mini-batch sampled uniformly at random from the local dataset of the $k$-th sensor. At the end of the training round of communication, our proposed federated stacked LSTM returns a vector of predicted values. Then, we concatenate all the predicted outcomes from the $K$ sensors to obtain a vector $\hat{\bm{y}}\in\mathbb{R}^{N}$  as follows:
\begin{equation}
\bm{\hat{y}} = \bm{\hat{y}}^1 \oplus \ldots \oplus \bm{\hat{y}}^K,
\end{equation}
where $\oplus$ denotes the concatenation operator.

Based on these predictions, the model sends back answers to the anomaly detection engine, where the values are passed on to the threshold determinator to make decisions regarding anomalous sensors, and subsequently take appropriate actions for isolation and maintenance.

\begin{algorithm}
  \caption{Federated Stacked LSTM}
  \label{alg:FSLSTM}
  \begin{algorithmic}[1]
    \REQUIRE Training sets $\{(\bm{X}^{k},\bm{y}^k)\}_{k=1}^{K}$ from $K$ sensors, initial global model parameters $\bm{w}_{0}$, local minibatch size $B$, number of local epochs $E$, learning rate $\eta$, number of rounds $R$, number of sensors per round $m$.
    \ENSURE Vector $\bm{\hat{y}}$ of predicted values.
    \FOR{$r=1$ to $R$}
    \STATE Server randomly selects a subset $S_{r}$ of $m$ sensors.
    \STATE Server broadcasts $\bm{w}_{r}$ to the subset $S_{r}$
    \FOR{each sensor $k\in S_{r}$ \textbf{in parallel}}
    \STATE $\bm{w}_{r+1}^{k} \leftarrow$ SensorUpdate$(k, \bm{w}_{r})$
    \ENDFOR

     $\bm{w}_{r+1} \leftarrow \sum_{k=1}^{K} \frac{N_k}{N}\bm{w}_{r+1}^{k}$     \hspace*{\fill}//\textit{aggregate updates}
    \ENDFOR
    \bigskip
  	\textbf{SensorUpdate}$(k, \bm{w})$: \hspace*{\fill}//\textit{run on sensor $k$}\\
  	$\mathcal{B}\leftarrow$ partition local data $(\bm{X}^{k},\bm{y}^k)$ into batches of size $B$
  	\FOR{each local epoch from 1 to $E$}
	 \FOR{mini-batch sample $\bg{\xi}\in\mathcal{B}$}
	 \STATE $\bm{h}_{t}, \bm{c}_{t} = \text{LSTM}(\bm{h}_{t-1},\bm{x}_{t},\bm{c}_{t-1};\bm{w})$ \hspace*{\fill}//\textit{LSTM at time step} $t$
	 \STATE $\hat{\bm{y}}^{k} = \sigma(\bm{W}^{\text{FC}} \bm{h}_{\text{last}}+ \bm{b}^{\text{FC}})$ \hspace*{\fill}//\textit{predicted outcomes}
     \STATE $\bm{w} \leftarrow \bm{w} - \frac{\eta}{B}\sum_{\bg{\xi}\in\mathcal{B}}\nabla\ell(\bm{w};\bg{\xi})$ \hspace*{\fill}//\textit{update local model}
	 \ENDFOR
  	\ENDFOR
    return learned parameter vector $\bm{w}$ to server
    \STATE Concatenate predicted values: $\bm{\hat{y}} \leftarrow \bm{\hat{y}}^1 \oplus \ldots \oplus \bm{\hat{y}}^K$
  \end{algorithmic}
\end{algorithm}

\section{Experiments}\label{sec:experiments}
In this section, we conduct experiments to demonstrate and analyze the performance of the proposed FSLSTM model in anomaly detection and regression on three real-world datasets generated by the IoT production system at General Electric Current smart building in Montreal. For privacy concerns, we hashed the primary keys and a few privacy-sensitive features such as vendor name, device identification and some of the prototypes information. The effectiveness of our approach is validated by performing comparison with several baseline methods.

\medskip\noindent{\textbf{Datasets.}}\quad We use a Sensors Event Log dataset for anomaly detection and an Energy Usage dataset for electricity consumption. We also use a Weather API dataset in conjunction with energy usage to enrich model learning. The weather data are collected from the building API that displays the basic weather parameters such as temperature, barometric pressure, humidity, precipitation, solar radiation and wind speed from May to August 2019. The weather data are used as a secondary source of information to enrich the Sensors Event Log dataset and help the model predict the energy usage. We simulate each zone in a separate category and allocate a specific task to it for learning.

\begin{figure}[!htb]
\centering
\includegraphics[scale=0.8]{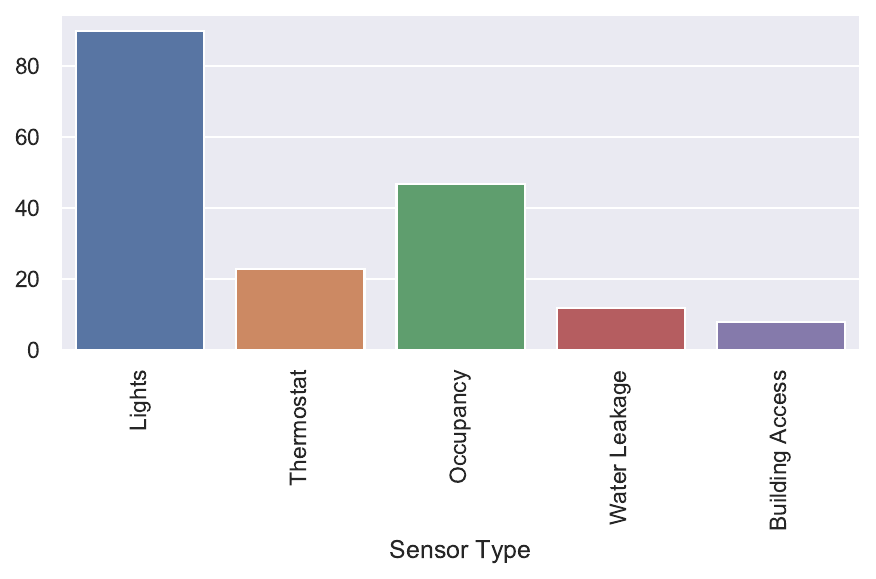}
\caption{Distribution of sensors' categories.}
\label{Fig:sensors_distribution}
\end{figure}

The Sensors Event Log and Energy Usage datasets are generated by 180 sensors from 5 different categories: lights, HVAC thermostats, occupancy sensors, water leakage sensors and access sensors. The distribution of these categories is shown in Figure \ref{Fig:sensors_distribution}. We use these categories for classification tasks, as the target is to identify which sensor is faulty, rendering it unable to properly communicate with HVAC and other control devices. In addition, the energy usage metric of thermostat sensors is used for regression tasks, with the aim of predicting future energy consumption of IoT devices, such as temperature, humidity and pressure sensors which are the most commonly used sensors for HVAC and building equipment applications. The pre-processing phase of data generation and wrangling are performed to ensure the confidentiality of the datasets used in our experiments and the privacy of the company's infrastructure.

\begin{itemize}
  \item\textbf{Sensors Event Log Dataset:} Sensors data such as occupancy sensors, lights, thermometer and humidity are collected from 180 devices. These sensors are categorized into 5 different groups and are distributed all over the building. In our experiments, we select 1 million event logs with a window time of 4 months. We limit data collection to one particular season (summer in our case) to learn feature vectors of time and frequency domain variables for this specific period of the year. We believe that each season should be considered separately for better learning purposes. For example, the heating pattern during the winter season is different from the cooling stages during summer. We consider each sensor as a separate task and predict its activities such as drop in temperature, excessive energy usage and running water.
  \item\textbf{Energy Usage Dataset:} Energy data gathered from sensors measure the electricity consumed per device. In our case, appliances can represent any equipment that is connected to the building automation system and falls under the five aforementioned categories. The unit is measured in kW/h for a variety of appliances, including LED light bulbs, rooftop units, humidity sensors, smart sensors that capture water leakage events, and occupancy activities around the building. The latter are sophisticated sensors, which are equipped with a smart dashboard for data analytics and different measuring tools to monitor important indicators related to indoor farms and laboratories. The data are aggregated every 15 minutes, stored in the Energy Usage table, and then merged into the Event Log dataset based on the sensor ID.
\end{itemize}

\medskip
\noindent{\textbf{Baseline methods.}}\quad We evaluate the performance of the proposed FSLSTM network consisting of three LSTM layers against several baseline models, including centralized logistic regression (LR), LSTM, federated logistic regression (FLR), federated gated recurrent unit (FGRU), federated LSTM with one LSTM layer (FLSTM-1), and federated LSTM with two LSTM layers (FLSTM-2). Both LSTM and GRU networks address the vanishing/exploding gradient problem of traditional recurrent neural networks. While the LSTM block consists of three gates (forget, input and output), the GRU block has only two gates (update and reset). The update gate in the GRU block determines the amount of previous information that needs to be passed along the next state, while the reset gate decides how much of the past information is needed to neglect. It is worth pointing out that we designed the federated logistic regression and federated GRU baseline models for comparison purposes with FSLSTM, whereas the centralized logistic regression and LSTM are well-known prediction baselines in the literature. Our proposed framework leverages the popular federated averaging algorithm, which uses stochastic gradient descent as both sensor and server optimizers to update the local and global parameters, where the server learning rate is equal to 1. More recently, Sashank \textit{et al.}~\cite{Sashank:2021} have introduced an adaptive optimization framework, in which federated versions of popular adaptive algorithms are incorporated for the sensor-side or server-side model updates.

\medskip
\noindent{\textbf{Implementation details.}}\quad All experiments are carried out on a Linux desktop computer running 4.4GHz and 64GB RAM with an NVIDIA GeForce RTX 2080 Ti GPU. The algorithms are implemented in PyTorch. The hyper-parameters are optimized using grid search. For fair comparison, we set the number of hidden units to 128 per layer in both LSTM and GRU blocks. We also set the number of nodes in the fully connected layer to 100. In addition, we use regular and recurrent dropouts of 20\% and apply the ReLU activation function in an effort to avoid overfitting. Our grid search selects a batch size of 1024 for training and an initial learning rate of 0.001 as the best combination. We use cross-entropy and MSE as loss functions for classification and regression tasks, respectively. In all experiments, we split our datasets into 80\% training and 20\% testing.

\subsection{Results}
The effectiveness of our FSLSTM model (i.e. FLSTM-3) is assessed by conducting a comprehensive comparison with the baseline methods using several performance evaluation metrics. The results are summarized in Table~\ref{Tab:Comparative_results_all}.

\begin{table}[!htb]
\caption{Performance comparison results of FSLSTM against baselines models on the Sensors Event Log test set for anomaly detection (classification) and Energy Usage test set (regression). Boldface numbers indicate the best performance.}
\medskip
\centering
\begin{tabular}{l*{8}{c}}
\toprule
\multicolumn{7}{c}{Sensors Data} &\multirow{1}{*}{}{EU Data}\\
\cmidrule(r){2-5} \cmidrule(r){7-9}
\text{Methods} &  \text{Precision} & \text{Recall} & \text{F1} & \text{BA} & & \text{} \text{MAE}& \text{MSE} & \text{RMSE}\\
\midrule
LR & 0.57 & 0.60 & 0.52 & 0.72 & \text{} & 0.341 & 0.48 & 0.692 \\
LSTM & 0.66 & 0.61 & 0.58 & 0.71 & \text{} & 0.243 & 0.33 & 0.574\\
FLR (ours) & 0.65 & 0.71 & 0.70 & 0.69 & \text{}& 0.339 & 0.34 & 0.583\\
FGRU (ours) & 0.84& 0.66 & 0.59 & 0.80 & \text{}& 0.211 & 0.29 & 0.538\\
FSLSTM (ours) & \textbf{0.89}&\textbf{0.79}&\textbf{0.87}&\textbf{0.90}& & \textbf{0.162} & \textbf{0.19} & \textbf{0.435}\\
\bottomrule
\end{tabular}
\label{Tab:Comparative_results_all}
\end{table}

For regression tasks, we use the mean absolute error (MAE), mean squared error (MSE) and root mean squared error (RMSE) as evaluation metrics, which are given by
\begin{equation}
\text{MAE} = \frac{1}{m}\sum_{i=1}^{m}|y_i - \hat{y}_i|,
\label{Eq:MAE}
\end{equation}
\begin{equation}
\text{MSE} = \frac{1}{m}\sum_{i=1}^{m}(y_i - \hat{y}_i)^2,
\end{equation}
and $\text{RMSE} = \sqrt{\text{MSE}}$, where $m$ is the number of samples in the test set, $y_i$ is the actual (ground truth) value, and $\hat{y}_i$ is the model's predicted value. A small value of these error metrics indicates a better performance of the model. We rely on a 15 minute window aggregation to reduce buffer overflow over the network and also to remove missing values that are sometimes generated by some sensors due to inactivity. As shown in Table~\ref{Tab:Comparative_results_all}, FSLSTM significantly outperforms the baselines methods on the Energy Usage datasets. In terms of MAE, for example, FSLSTM yields a much smaller error compared to the one obtained by LSTM, and more than half the error obtained by LR and FLR. Moreover, FSLSTM yields a 4.9 percentage points performance improvement over FGRU.

For classification tasks, we use precision, recall, F1 score, balanced accuracy (BA), receiver operating characteristic (ROC) curve and area under the ROC curve (AUC) as evaluation metrics. Precision and recall are defined as
\begin{equation}
\text{Precision} = \frac{\text{TP}}{\text{TP}+\text{FP}},\quad\text{and}\quad \text{Recall} = \frac{\text{TP}}{\text{TP}+\text{FN}},
\end{equation}
where TP, FP, TN and FN denote true positives, false positives, true negatives and false negatives, respectively. TP is the number of correctly predicted anomalous observations, while TN is the number of correctly predicted normal observations. Recall, also known as true positive rate (TPR), is the percentage of positive instances correctly classified, and indicates how often a classifier misses a positive prediction.

AUC summarizes the information contained in the ROC curve, which plots TPR versus $\text{FPR}=\text{FP}/(\text{FP}+\text{TN})$, the false positive rate, at various thresholds. Larger AUC values indicate better performance at distinguishing between anomalous and normal observations. FPR is is the ratio of normal observations that were incorrectly classified as anomalous.

Since our datasets are highly imbalanced, we use the balanced F1 score defined as the harmonic mean of precision and recall, and the balanced accuracy given by
\begin{equation}
\text{BA} = \frac{\text{TPR} + \text{TNR}}{2},
\end{equation}
where $\text{TNR}=1-\text{FPR}$, also called specificity, refers to the ability of a classifier to correctly identify the observations that are not anomalous.

As shown in Table~\ref{Tab:Comparative_results_all}, the proposed model outperforms all the baseline methods by a large margin in terms of F1 score, yielding performance improvements of 29 percentage points over LSTM and 28 percentage points over FGRU. In terms of balanced accuracy, FSLSTM also achieves notable performance improvements of 19 percentage points over LSTM and 10 percentage points over FGRU. This demonstrates the significant prediction ability of FSLSTM in anomaly detection. It is also important to note that LSTM yields comparable performance to FLR, due in large part to the fact that linear models cannot effectively represent temporal dependency, even in a federated learning setting.

\begin{figure}[!htb]
\centering
\includegraphics[scale=0.8]{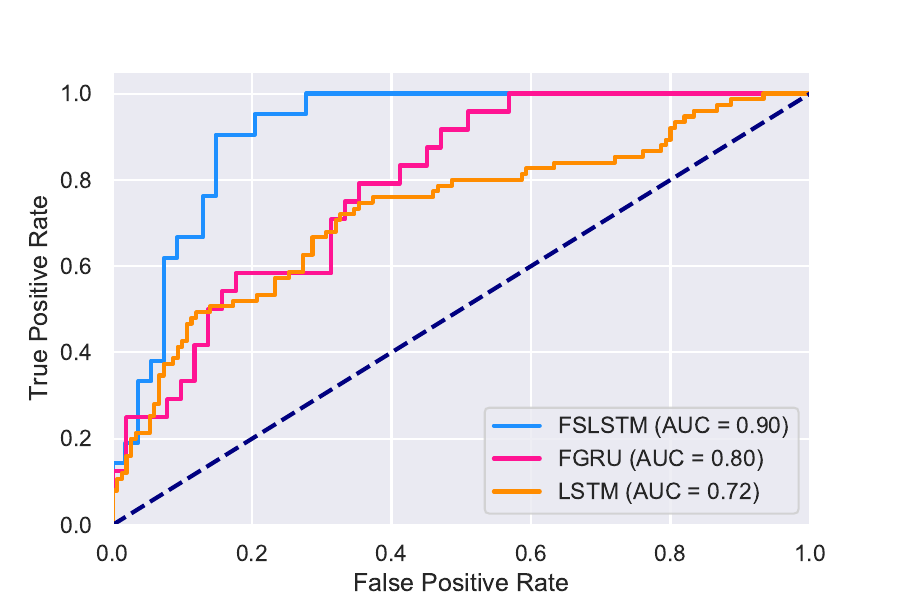}
\caption{ROC curves for FSLSTM and baseline methods, along with the corresponding AUC values, on the Sensors Event Log test set.}
\label{Fig:roc_curve}
\end{figure}

Figure~\ref{Fig:roc_curve} displays the ROC curves, which show the better performance of FSLSTM in comparison with baseline methods on the Sensors Event Log test set. Each point on ROC represents different trade-off between false positives and false negatives. An ROC curve that is closer to the upper right indicates a better performance (TPR is higher than FPR). The overall performance of FSLSTM is significantly better than the baselines, as indicated by both the ROC curves and AUC values.

\medskip
\noindent{\textbf{Federated vs. centralized learning.}}\quad The LSTM model adopts a centralized approach, which requires the training data to be aggregated on the server, meaning that no training is performed on the edge devices. The learning is carried out by partitioning the data into training and test sets in a central machine. However, centralized training is privacy-intrusive, especially for clients with important sensors' data aggregated on the cloud, as some sensors may contain private patterns, such as occupancy and access information. In a centralized setting, sensors have to trade their privacy by sending and storing data on the cloud owned by a third-party service provider.

Unlike the centralized training framework, our federated learning based approach is decentralized and enables sensors located at different locations inside the building to collaboratively learn a federated model, while keeping all data that may contain private information on devices. Therefore, sensors can benefit from using a federated model without sending raw data to the cloud. Sensitive data may include the sensor's model number, manufacturer name or even serial numbers of newly produced prototypes that are still in the testing phase. Another limitation of centralized learning is the computing power~\cite{chen2015efficient}. In our setting, however, the computing power for each sensor is negligible compared to dedicated GPUs. In a centralized learning environment, a large amount of data collected from different IoT devices need to be merged into one dataset and then we wait for the machine to finish training. In our proposed federated framework, we train the model on the data of each device in parallel, resulting in better performance and faster convergence time. The better performance of our federated stacked LSTM model may be attributed not only to the strong capabilities of LSTM in prediction tasks, but also to the stacked LSTM layers that help generate a deep feature representation of the input data.

\begin{figure}[!htb]
\centering
\includegraphics[scale=0.8]{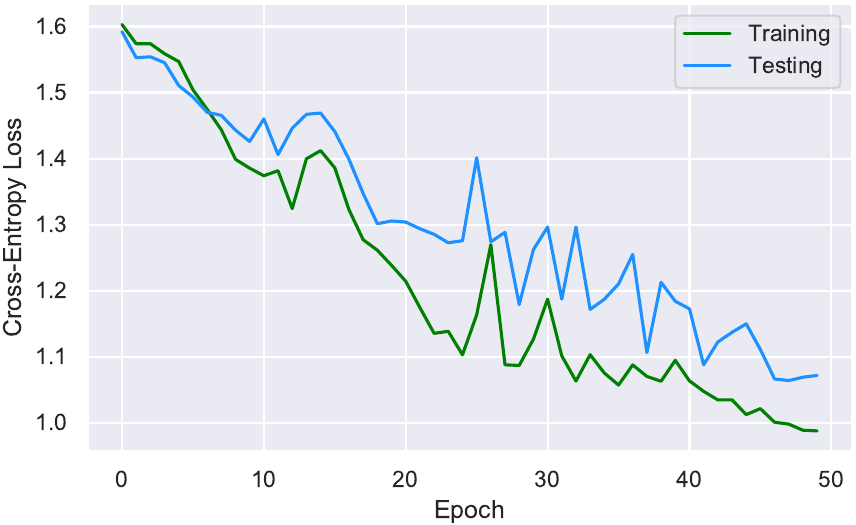}\\[1ex]
\includegraphics[scale=0.8]{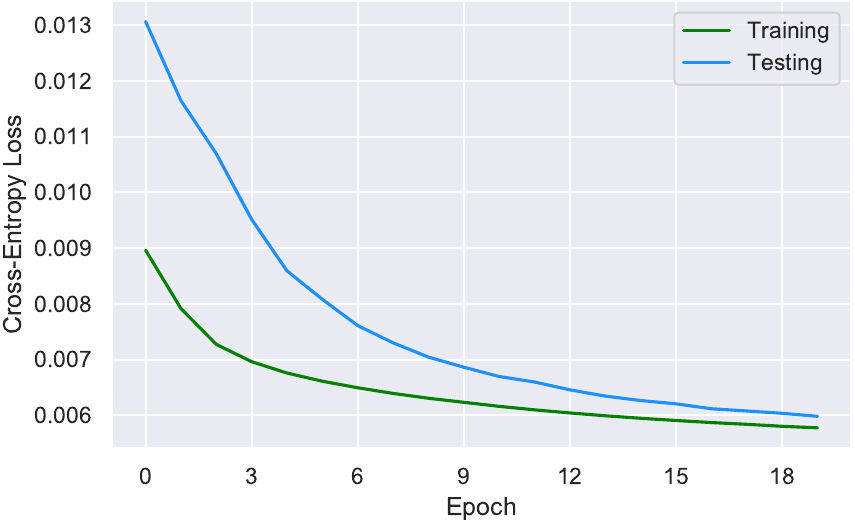}
\caption{Training and testing learning curves of LSTM (top) and FSLSTM (bottom) on the Sensors Event Log dataset.}
\label{Fig:FLSTM}
\end{figure}

The ability to quickly detect and respond to anomalies is critical to the successful operation of smart buildings. Models trained on data of individual devices help improve critical incidents detection, such as technical glitches that may arise in smart buildings, where smart fire sensors, for instance, need not only be triggered in the case of an emergency, but also be capable of predicting specific scenarios and patterns. During the training phase, our FSLSTM model converges twice as fast as the centralized LSTM on the same datasets. As shown in Figure~\ref{Fig:FLSTM} (top), the centralized LSTM model does not seem to attain a stable state even after 50 epochs. The federated stacked LSTM model, on the other hand, is capable of significantly reducing the loss function and reaching a stable performance with higher accuracy in only 20 epochs, as shown in Figure~\ref{Fig:FLSTM} (bottom). Each epoch represents a full round of the 180 sensors that are participating in the experiment. Moreover, notice that the learning curves of the federated model are less fluctuating compared to the centralized LSTM. This smoothness property is attributed, in large part, to the number of sensors involved in the learning process.

\begin{figure}[!htp]
\centering
\includegraphics[scale=0.8]{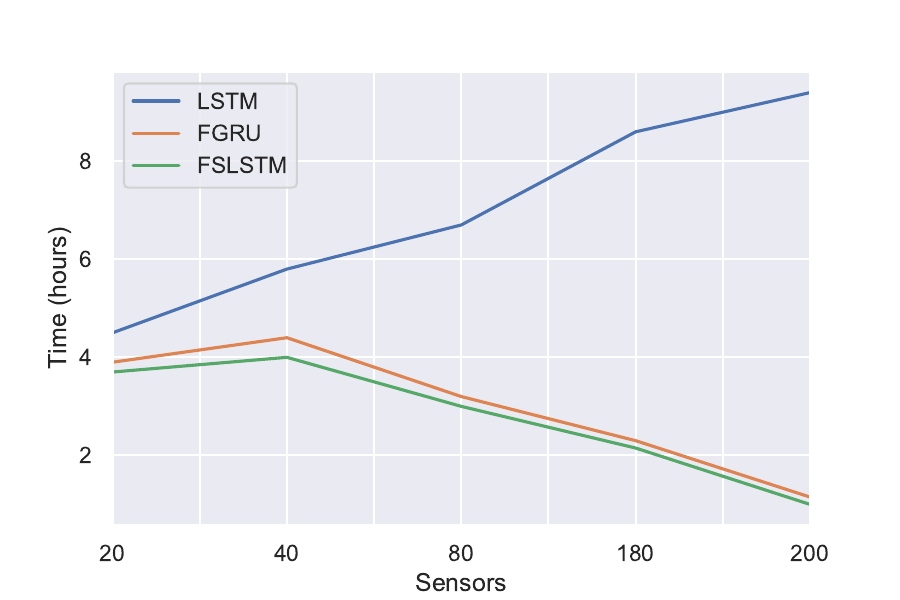}
\caption{Convergence time comparison between FSLSTM, FGRU and LSTM using a varying number of sensors on the Sensors Event Log dataset.}
\label{Fig:Convergence_Speed}
\end{figure}

\medskip
\noindent{\textbf{Convergence performance.}}\quad An important factor in our experiments is the convergence speed of the model. We test the effect of the parameter $K$ (i.e. number of sensors) on the convergence performance of the proposed approach. The results of convergence time for FSLSTM, FGRU and LSTM using a varying number of sensors on the Sensors Event Log dataset are shown in Figure~\ref{Fig:Convergence_Speed}. Note that unlike the centralized LSTM model, the convergence time of both federated models decreases when the number of sensors increases. Training a deep neural network involves using an optimization algorithm to find a set of weights to best map inputs to outputs, while convergence describes a progression towards a stable state where the network has learned to properly respond to a set of training patterns within some margin of error. The choice of the network's hyperparameters, such as the number of sensors, play an important role in convergence. In order to assess the performance of FSLSTM with respect to the number of sensors $K$, we use the mini-batch stochastic gradient descent optimization algorithm with a fixed batch size for the local update by increasing the value of $K$ from 20 to 200. For all sensors, even when the learning rate is tuned carefully, FSLSTM achieves, in all batch size settings, a better performance, as reported in Table~\ref{Tab:Comparative_results_all}. Figure~\ref{Fig:Convergence_Speed} shows that an increase in the number of participating sensors positively impacts the convergence time of FSLSTM, reducing the time from 6 hours for the centralized LSTM to only 2 hours for FSLSTM using 200 sensors. However, a larger number of participating sensors requires higher local computation at the sensors, resulting in an increase in energy consumption by these sensors.

It is also important to point out that there a is slight increase of convergence time when the number of sensors is between 20 and 40, but then the convergence time decreases when the number of sensors exceeds 42. This is largely due to the delay in response of the rooftop units' thermostats that are connected to the network through the Wireless Area Controller gateways. The latency time of these sensors is twice longer than other sensors in the buildings, leading to a slight increase in convergence time. Once these sensors are equipped with all the global parameters that are necessary for the training phase, the convergence time drops proportionally to the number of participating sensors.

\medskip
\noindent{\textbf{Collective and contextual anomalies.}} \quad A collective anomaly describes a group of data points that exhibits an anomalous behavior compared to the rest of the dataset~\cite{zheng2015detecting}. An individual instance within the anomalous group is not necessarily anomalous on its own. If a data instance is anomalous in a specific context, but not otherwise, then it is termed as a contextual anomaly, also referred to as conditional anomaly~\cite{hayes2015contextual}. We train FSLSTM on normal data before performing a live prediction for each time step, which is equal to 600 minutes. Instead of considering each time step separately, the observation of prediction errors from a certain number of time steps is now used for detecting collective anomalies. The prediction errors from a number of the latest time steps above a threshold, as set by the threshold determinator, will indicate a collective anomaly. We take a sample of our initial data, which includes four HVAC sensors, all measuring the electricity from power meters located in different areas within the building. In the first step, FSLSTM determines the point anomalies in real-time, while in the second step, the anomaly detection engine decides if these anomalies are contextual or collective based on different alarm profiles and system rules, as well as specific temperature profiles for each category of devices. Our proposed model efficiently determines context/collection based anomalies in real-time, as reported in Table~\ref{Tab:colcont}.

As expected, we noticed through experimentation that it is possible to obtain a higher accuracy on collective anomalies detection, but the number of false alarms triggered tends to be quite high. As shown in in Table~\ref{Tab:colcont}, FSLSTM yields superior performance by correctly detecting 88\% of the alarms out of 1000 instances in this experiment with only 9 false alarms. On the contextual side, FSLSTM achieves even a better performance of 90\% with only 4 false alarms.

\begin{table}[!htb]
\caption{Collective and contextual anomaly detection comparison.}
\medskip
\centering
\begin{tabular}{lcccc}
\toprule
\multirow{2}[3]{*}{Method} & \multicolumn{2}{c}{Collective Anomalies} & \multicolumn{2}{c}{Contextual Anomalies} \\
\cmidrule(lr){2-3} \cmidrule(lr){4-5}
 & \text{Correct Alarms} & \text{False Alarms} & \text{Correct Alarms} & \text{False Alarms}\\
 & \text{Triggered (\%)} & \text{Triggered (\%)}& \text{Triggered (\%)} & \text{Triggered (\%)}\\
\midrule
\text{LR} & 56 & 54  & 63 & 48\\
\text{LSTM} & 66 & 33 & 74 & 29\\
\text{FLR (ours)} & 65 & 21 & 78 & 18\\
\text{FGRU (ours)} & 74 & 12 & 82 & 7\\
\text{FSLSTM (ours)} & \textbf{88} & \textbf{9} & \textbf{90} & \textbf{4}\\
\bottomrule
\end{tabular}
\label{Tab:colcont}
\end{table}

\begin{figure}[!htb]
\centering
\includegraphics[scale=0.34]{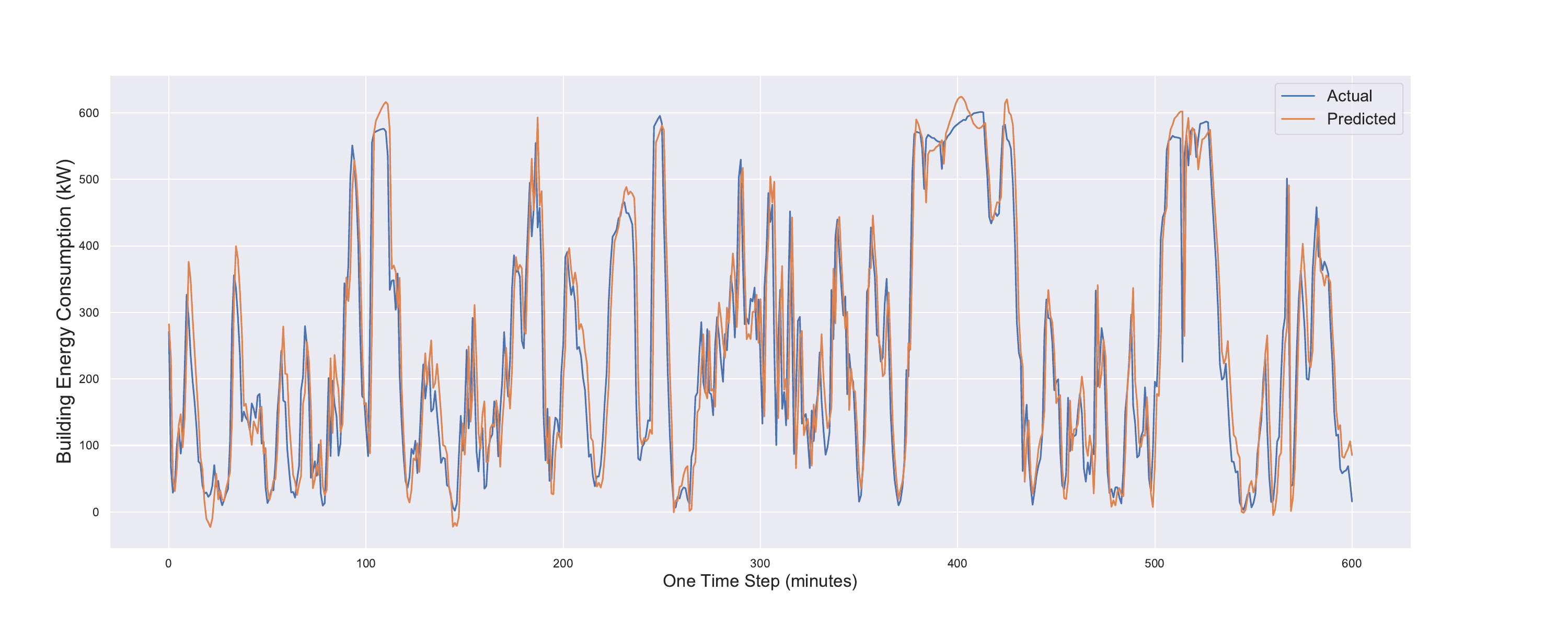}
\caption{Actual vs. predicted building energy consumption using FSLSTM on the Energy Usage test set.}
\label{Fig:actual_predicted}
\end{figure}

\medskip
\noindent{\textbf{Prediction performance.}}\quad For regression tasks, we test the performance of FSLSTM on the Energy Usage dataset to predict the building energy consumption on a window time of 600 minutes for two main reasons. First, the confidence interval of the model shows a high accuracy of energy prediction, as illustrated in Figure~\ref{Fig:actual_predicted}, which displays the actual vs. predicted building energy consumption using FSLSTM. Second, it makes more sense from an industrial perspective as 600 min equals 10 hours, which roughly represents a full working day. This time frame provides building managers with plenty of insightful information to anticipate and predict the daily usage of energy, and allows them to plan activities accordingly. As shown in Figure~\ref{Fig:actual_predicted}, the FSLSTM model has a stable performance and is able to predict energy consumption in the building with 90\% accuracy.

\begin{figure}[!htb]
\setlength{\tabcolsep}{.05em}
\centering
\begin{tabular} {cccc}
\includegraphics[scale=.24]{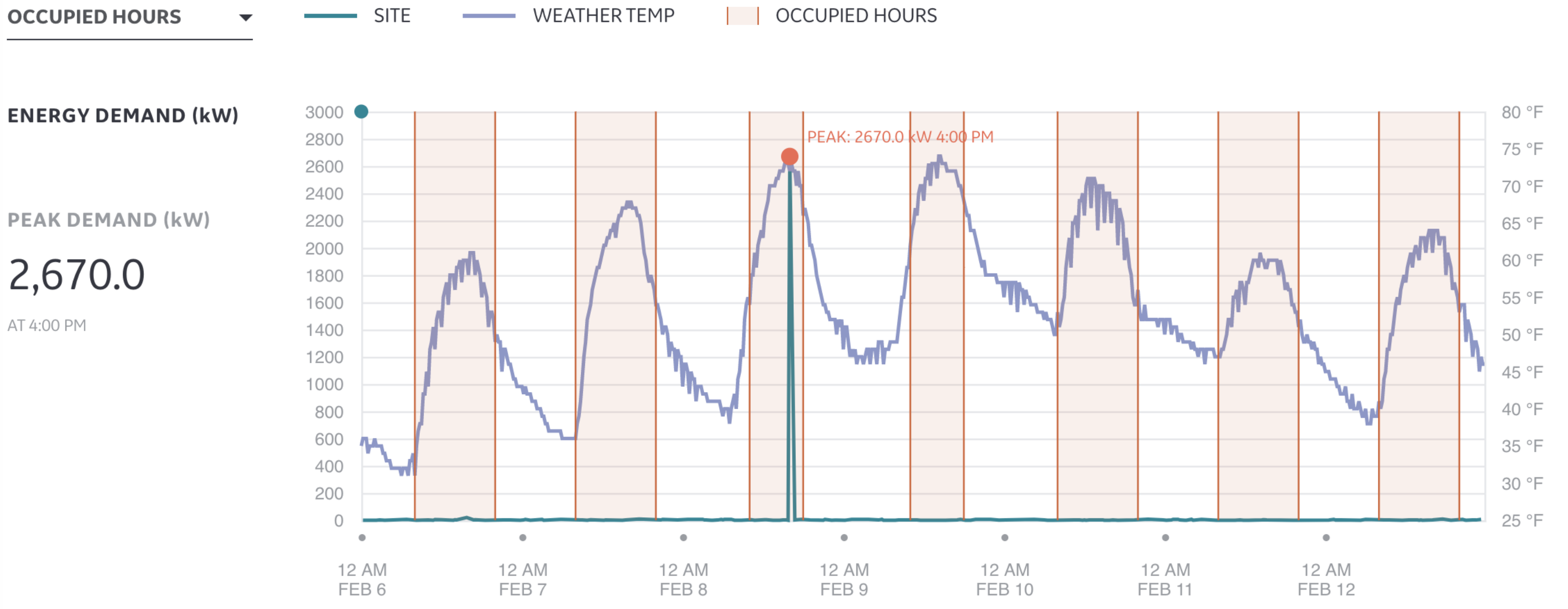}&\\
\includegraphics[scale=.24]{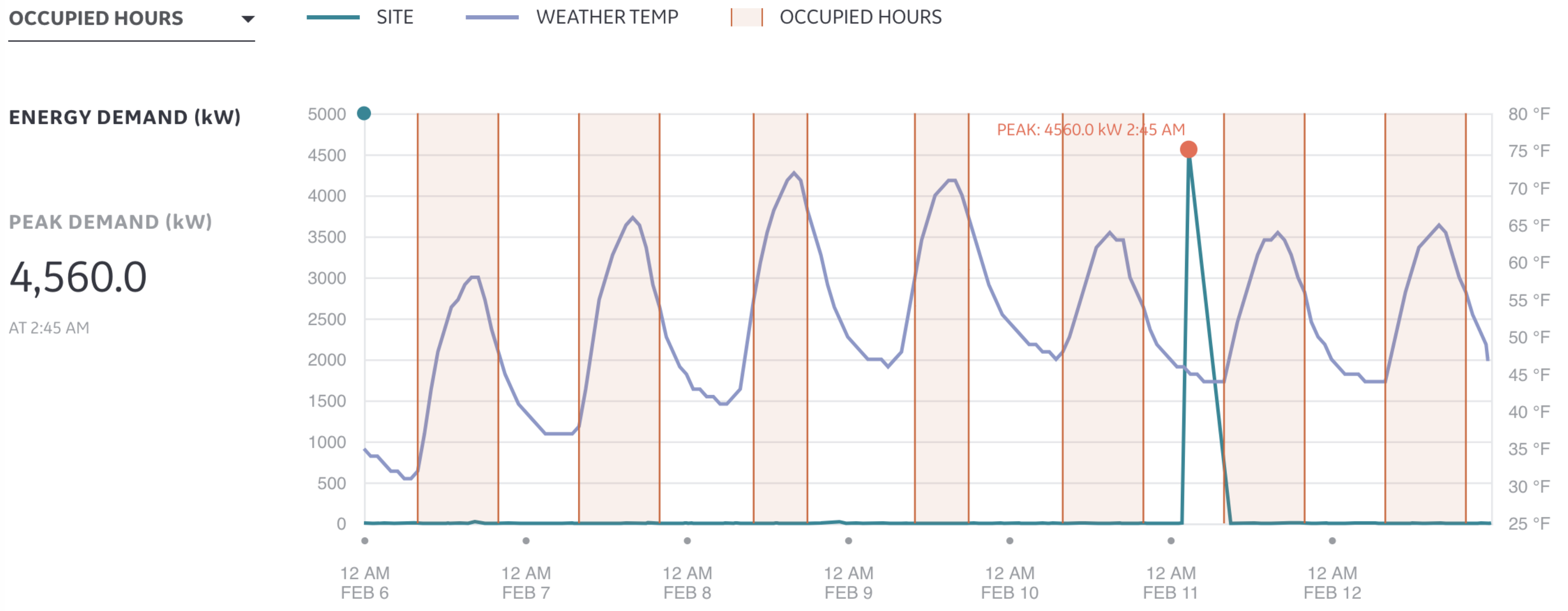} &\\
\includegraphics[scale=.24]{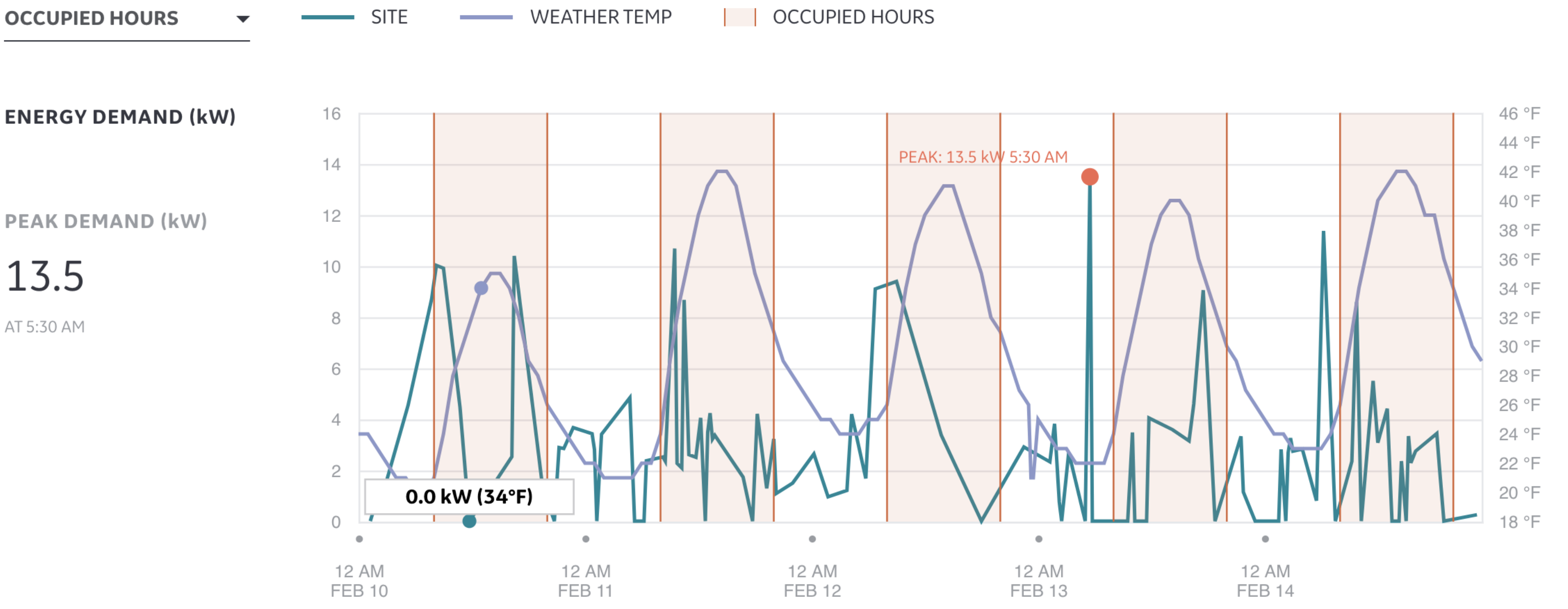}
\end{tabular}
\caption{Testing FSLSTM for anomaly detection on three types of sensors: lights, thermostats and water leakage. The red dot indicates an anomaly.}
 \label{Fig:isolines}
\end{figure}

For the sensors fault detection, we run the classification task on the Sensors dataset to detect anomalous devices. The experiment includes lights, thermostats of rooftop units and water leakage sensors for a window time ranging from 4 to 6 days. Figure~\ref{Fig:isolines} demonstrates the effectiveness of the FSLTM model in detecting outstanding and malfunctioning behaviors expressed by these sensors at different times of the day. As can be seen, the model is able to capture outliers with high accuracy, along with the corresponding value and time stamp. During working hours (light orange color), the weather temperature (dark gray color) shows an increase during the day due to sunlight and heat, while the site energy demand (green color) is stable during the majority of hours. As shown in Figure~\ref{Fig:isolines} (top), the weather temperature is higher during working hours than during non-working hours (white bars). During this window time, the site has stable energy demands ranging between 0 and 35 Kw. FSLSTM is capable of capturing an anomalous amount of energy demands, caused by a cluster of malfunctioning devices, highlighting it as an outlier in real-time and eventually recording these devices with their manufacturing information and sending it back to the federated learning system. Then, the threshold determinator makes the appropriate decision and sends it over the BAS system for maintenance.

\begin{figure}[!htb]
\centering
\includegraphics[scale=0.75]{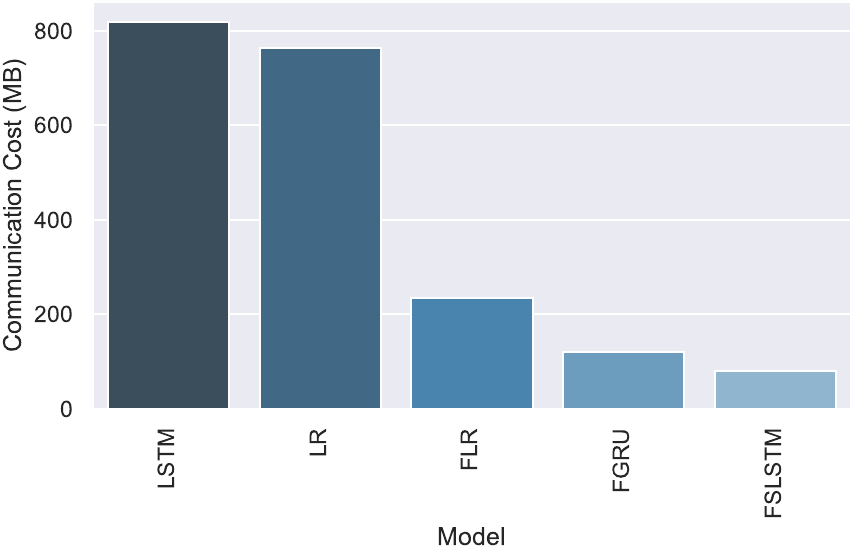}
\caption{Communication cost comparison between FSLSTM and baseline methods, all trained for 50 epochs.}
\label{Fig:communication_cost}
\end{figure}

\medskip
\noindent{\textbf{Communication Cost}}\quad One of the key challenges in our federated learning system is the communication cost, which is typically expressed as a function of data volume, e.g. Megabytes. As shown in Figure~\ref{Fig:communication_cost}, the proposed FSLSTM model significantly reduces the communication cost compared to the baseline methods on the Sensors Event Log dataset. We argue that the superior performance of FSLSTM over the centralized LSTM and LR models is largely attributed to the effective participation of sensors in the training phase~\cite{yao2018two}. Federated learning models enable decentralized IoT devices to collaboratively learn a shared prediction model without sending the actual data that is generated during the specific round to the central server. This property of federated learning enhances the communication quality and reduces the cost, while keeping all the training data on device and hence reducing concerns on data security and privacy.

\subsection{Ablation Study}
In order to validate the effectiveness of our FSLSTM framework, we perform an ablation study under different configurations by changing the number of LSTM layers and retraining the models on the same datasets. We design the FLSTM-1 and FSLSTM-2 models by removing one and two LSTM layers, respectively, from the proposed FLSTM architecture, which consists of three LSTM layers. The results are reported in Table~\ref{Tab:Comparative_fed}, which shows that both FSLSTM-1 and FSLSTM-2 achieve better performance than LSTM and FGRU. It is important to note that with only a single LSTM layer, the federated LSTM model is able to outperform all baselines methods in terms of balanced accuracy and mean absolute error metrics. We also experimented with more than three LSTM layers, but we did not notice significant performance improvements. In addition, some small sensors (e.g. water leakage sensors) are unable to support the growing size of the model.

\begin{table}[!htb]
\caption{Ablation study of FSLSTM on the Sensors Event Log and Energy Usage test sets using different LSTM layers. Boldface numbers indicate the best performance.}
\medskip
\centering
\begin{tabular}{l*{8}{c}}
\toprule
\multicolumn{7}{c}{Sensors Data} &\multirow{1}{*}{}{EU Data}\\
\cmidrule(r){2-5} \cmidrule(r){7-9}
\text{Methods} &  \text{Precision} & \text{Recall} & \text{F1} & \text{BA} & & \text{} \text{MAE}& \text{MSE} & \text{RMSE}\\
\midrule
\text{FLSTM-1}&0.85& 0.68 & 0.69 & 0.81 &\textbf{} & 0.192 & 0.24 & 0.489\\
\text{FLSTM-2}&0.87 & 0.73 & 0.71 & 0.83&\textbf{} & 0.171 & 0.21 & 0.458\\
\text{FSLSTM}&\textbf{0.89}&\textbf{0.79}&\textbf{0.87}&\textbf{0.90}& & \textbf{0.162} & \textbf{0.19} & \textbf{0.435}\\
\bottomrule
\end{tabular}
\label{Tab:Comparative_fed}
\end{table}

\section{Conclusion}
In this paper, we introduced a federated stacked LSTM framework for anomaly detection in smart buildings using federated learning for IoT sensor data. The proposed FSLSTM network consists of a local LSTM model that captures individual sensors' data and a global model that aggregates the weights, updates the parameters and shares them again with sensors across all tasks. Experimental results on two datasets demonstrated FLSTM's ability to significantly improve the results of a variety of centralized models in IoT settings, achieving much better performance compared to baseline methods in both classification and regression tasks. We also showed that our federated stacked LSTM model converges twice as fast than the centralized LSTM during the training phase. In addition, we conducted an ablation study on our FSLSTM network to evaluate its performance and robustness under different configurations by changing the number of LSTM layers and retraining the models on the same datasets. In the future, we plan to investigate other IoT applications in a federated learning setting such as blockchain and electric vehicle charging networks.

\section*{Acknowledgments}
This work was supported in part by the Natural Sciences and Engineering Research Council of Canada through the Discovery Grants Program.

\bibliographystyle{ACM-Reference-Format}
\bibliography{references}


\begin{thebibliography}{47}


\ifx \showCODEN    \undefined \def \showCODEN     #1{\unskip}     \fi
\ifx \showDOI      \undefined \def \showDOI       #1{#1}\fi
\ifx \showISBNx    \undefined \def \showISBNx     #1{\unskip}     \fi
\ifx \showISBNxiii \undefined \def \showISBNxiii  #1{\unskip}     \fi
\ifx \showISSN     \undefined \def \showISSN      #1{\unskip}     \fi
\ifx \showLCCN     \undefined \def \showLCCN      #1{\unskip}     \fi
\ifx \shownote     \undefined \def \shownote      #1{#1}          \fi
\ifx \showarticletitle \undefined \def \showarticletitle #1{#1}   \fi
\ifx \showURL      \undefined \def \showURL       {\relax}        \fi
\providecommand\bibfield[2]{#2}
\providecommand\bibinfo[2]{#2}
\providecommand\natexlab[1]{#1}
\providecommand\showeprint[2][]{arXiv:#2}

\bibitem[\protect\citeauthoryear{Bellido-Outeirino, Flores-Arias,
  Domingo-Perez, Gil-de Castro, and Moreno-Munoz}{Bellido-Outeirino
  et~al\mbox{.}}{2012}]%
        {bellido2012building}
\bibfield{author}{\bibinfo{person}{Francisco~Jose Bellido-Outeirino},
  \bibinfo{person}{Jose~Maria Flores-Arias}, \bibinfo{person}{Francisco
  Domingo-Perez}, \bibinfo{person}{Aurora Gil-de Castro}, {and}
  \bibinfo{person}{Antonio Moreno-Munoz}.} \bibinfo{year}{2012}\natexlab{}.
\newblock \showarticletitle{Building lighting automation through the
  integration of {DALI} with wireless sensor networks}.
\newblock \bibinfo{journal}{\emph{IEEE Transactions on Consumer Electronics}}
  \bibinfo{volume}{58}, \bibinfo{number}{1} (\bibinfo{year}{2012}),
  \bibinfo{pages}{47--52}.
\newblock


\bibitem[\protect\citeauthoryear{Beltran and Cerpa}{Beltran and Cerpa}{2014}]%
        {beltran2014optimal}
\bibfield{author}{\bibinfo{person}{Alex Beltran} {and}
  \bibinfo{person}{Alberto~E Cerpa}.} \bibinfo{year}{2014}\natexlab{}.
\newblock \showarticletitle{Optimal HVAC building control with occupancy
  prediction}. In \bibinfo{booktitle}{\emph{Proc. ACM Conference on Embedded
  Systems for Energy-Efficient Buildings}}. \bibinfo{pages}{168--171}.
\newblock


\bibitem[\protect\citeauthoryear{Bonawitz, Eichner, Grieskamp, Huba, Ingerman,
  Ivanov, Kiddon, Konecn\'y, Mazzocchi, McMahan, Overveldt, Petrou, Ramage, and
  Roselander}{Bonawitz et~al\mbox{.}}{2019}]%
        {KBonawitz:19}
\bibfield{author}{\bibinfo{person}{Keith Bonawitz}, \bibinfo{person}{Hubert
  Eichner}, \bibinfo{person}{Wolfgang Grieskamp}, \bibinfo{person}{Dzmitry
  Huba}, \bibinfo{person}{Alex Ingerman}, \bibinfo{person}{Vladimir Ivanov},
  \bibinfo{person}{Chloe Kiddon}, \bibinfo{person}{Jakub Konecn\'y},
  \bibinfo{person}{Stefano Mazzocchi}, \bibinfo{person}{H.~Brendan McMahan},
  \bibinfo{person}{Timon~Van Overveldt}, \bibinfo{person}{David Petrou},
  \bibinfo{person}{Daniel Ramage}, {and} \bibinfo{person}{Jason Roselander}.}
  \bibinfo{year}{2019}\natexlab{}.
\newblock \showarticletitle{Towards Federated Learning at Scale: System
  Design}. In \bibinfo{booktitle}{\emph{Conference on Systems and Machine
  Learning}}.
\newblock


\bibitem[\protect\citeauthoryear{Bonawitz, Ivanov, Kreuter, Marcedone, McMahan,
  Patel, Ramage, Segal, and Seth}{Bonawitz et~al\mbox{.}}{2017}]%
        {bonawitz2017practical}
\bibfield{author}{\bibinfo{person}{Keith Bonawitz}, \bibinfo{person}{Vladimir
  Ivanov}, \bibinfo{person}{Ben Kreuter}, \bibinfo{person}{Antonio Marcedone},
  \bibinfo{person}{H~Brendan McMahan}, \bibinfo{person}{Sarvar Patel},
  \bibinfo{person}{Daniel Ramage}, \bibinfo{person}{Aaron Segal}, {and}
  \bibinfo{person}{Karn Seth}.} \bibinfo{year}{2017}\natexlab{}.
\newblock \showarticletitle{Practical secure aggregation for privacy-preserving
  machine learning}. In \bibinfo{booktitle}{\emph{Proc. ACM Conference on
  Computer and Communications Security}}. \bibinfo{pages}{1175--1191}.
\newblock


\bibitem[\protect\citeauthoryear{Chandra and Cripps}{Chandra and
  Cripps}{2018}]%
        {chandra2018bayesian}
\bibfield{author}{\bibinfo{person}{Rohitash Chandra} {and}
  \bibinfo{person}{Sally Cripps}.} \bibinfo{year}{2018}\natexlab{}.
\newblock \showarticletitle{Bayesian multi-task learning for dynamic time
  series prediction}. In \bibinfo{booktitle}{\emph{Proc. IEEE International
  Joint Conference on Neural Networks}}. \bibinfo{pages}{1--8}.
\newblock


\bibitem[\protect\citeauthoryear{Chen, Qiu, Liu, and Huang}{Chen
  et~al\mbox{.}}{2018b}]%
        {chen2018meta}
\bibfield{author}{\bibinfo{person}{Junkun Chen}, \bibinfo{person}{Xipeng Qiu},
  \bibinfo{person}{Pengfei Liu}, {and} \bibinfo{person}{Xuanjing Huang}.}
  \bibinfo{year}{2018}\natexlab{b}.
\newblock \showarticletitle{Meta multi-task learning for sequence modeling}.
\newblock \bibinfo{journal}{\emph{Neurocomputing}} (\bibinfo{year}{2018}).
\newblock


\bibitem[\protect\citeauthoryear{Chen, Ho, Hsieh, Huang, Lee, and Mahajan}{Chen
  et~al\mbox{.}}{2018a}]%
        {chen2017adf}
\bibfield{author}{\bibinfo{person}{Ling-Jyh Chen}, \bibinfo{person}{Yao-Hua
  Ho}, \bibinfo{person}{Hsin-Hung Hsieh}, \bibinfo{person}{Shih-Ting Huang},
  \bibinfo{person}{Hu-Cheng Lee}, {and} \bibinfo{person}{Sachit Mahajan}.}
  \bibinfo{year}{2018}\natexlab{a}.
\newblock \showarticletitle{{ADF}: An anomaly detection framework for
  large-scale {PM}2.5 sensing systems}.
\newblock \bibinfo{journal}{\emph{IEEE Internet of Things Journal}}
  \bibinfo{volume}{5}, \bibinfo{number}{2} (\bibinfo{year}{2018}),
  \bibinfo{pages}{559--570}.
\newblock


\bibitem[\protect\citeauthoryear{Chen, Jiao, Li, and Fu}{Chen
  et~al\mbox{.}}{2015}]%
        {chen2015efficient}
\bibfield{author}{\bibinfo{person}{Xu Chen}, \bibinfo{person}{Lei Jiao},
  \bibinfo{person}{Wenzhong Li}, {and} \bibinfo{person}{Xiaoming Fu}.}
  \bibinfo{year}{2015}\natexlab{}.
\newblock \showarticletitle{Efficient multi-user computation offloading for
  mobile-edge cloud computing}.
\newblock \bibinfo{journal}{\emph{IEEE/ACM Transactions on Networking}}
  \bibinfo{volume}{24}, \bibinfo{number}{5} (\bibinfo{year}{2015}),
  \bibinfo{pages}{2795--2808}.
\newblock


\bibitem[\protect\citeauthoryear{Chen, Ning, Chai, and Rangwala}{Chen
  et~al\mbox{.}}{2019}]%
        {YChen:19}
\bibfield{author}{\bibinfo{person}{Y. Chen}, \bibinfo{person}{Y. Ning},
  \bibinfo{person}{Z. Chai}, {and} \bibinfo{person}{H. Rangwala}.}
  \bibinfo{year}{2019}\natexlab{}.
\newblock \showarticletitle{Federated Multi-task Hierarchical Attention Model
  for Sensor Analytics}.
\newblock \bibinfo{journal}{\emph{arXiv preprint arXiv:1905.05142}}
  (\bibinfo{year}{2019}).
\newblock


\bibitem[\protect\citeauthoryear{Cook, M{\i}s{\i}rl{\i}, and Fan}{Cook
  et~al\mbox{.}}{2019}]%
        {cook2019anomaly}
\bibfield{author}{\bibinfo{person}{Andrew Cook}, \bibinfo{person}{G{\"o}ksel
  M{\i}s{\i}rl{\i}}, {and} \bibinfo{person}{Zhong Fan}.}
  \bibinfo{year}{2019}\natexlab{}.
\newblock \showarticletitle{Anomaly detection for {I}o{T} time-series data: A
  survey}.
\newblock \bibinfo{journal}{\emph{IEEE Internet of Things Journal}}
  (\bibinfo{year}{2019}).
\newblock


\bibitem[\protect\citeauthoryear{D{'}Oca, Hong, and Langevin}{D{'}Oca
  et~al\mbox{.}}{2018}]%
        {d2018human}
\bibfield{author}{\bibinfo{person}{Simona D{'}Oca}, \bibinfo{person}{Tianzhen
  Hong}, {and} \bibinfo{person}{Jared Langevin}.}
  \bibinfo{year}{2018}\natexlab{}.
\newblock \showarticletitle{The human dimensions of energy use in buildings: A
  review}.
\newblock \bibinfo{journal}{\emph{Renewable and Sustainable Energy Reviews}}
  \bibinfo{volume}{81} (\bibinfo{year}{2018}), \bibinfo{pages}{731--742}.
\newblock


\bibitem[\protect\citeauthoryear{Du, Li, Zheng, and Srikumar}{Du
  et~al\mbox{.}}{2017}]%
        {MinDu:17}
\bibfield{author}{\bibinfo{person}{Min Du}, \bibinfo{person}{Feifei Li},
  \bibinfo{person}{Guineng Zheng}, {and} \bibinfo{person}{Vivek Srikumar}.}
  \bibinfo{year}{2017}\natexlab{}.
\newblock \showarticletitle{{DeepLog}: Anomaly detection and diagnosis from
  system logs through deep learning}. In \bibinfo{booktitle}{\emph{Proc. ACM
  SIGSAC Conference on Computer and Communications Security}}.
\newblock


\bibitem[\protect\citeauthoryear{Geiping, Bauermeister, Dr\"{o}ge, and
  Moeller}{Geiping et~al\mbox{.}}{2020}]%
        {Jonas2020Geiping}
\bibfield{author}{\bibinfo{person}{Jonas Geiping}, \bibinfo{person}{Hartmut
  Bauermeister}, \bibinfo{person}{Hannah Dr\"{o}ge}, {and}
  \bibinfo{person}{Michael Moeller}.} \bibinfo{year}{2020}\natexlab{}.
\newblock \showarticletitle{Inverting Gradients - How easy is it to break
  privacy in federated learning?}. In \bibinfo{booktitle}{\emph{Advances in
  Neural Information Processing Systems}}.
\newblock


\bibitem[\protect\citeauthoryear{Goodhue and Straub}{Goodhue and
  Straub}{1991}]%
        {goodhue1991security}
\bibfield{author}{\bibinfo{person}{Dale~L Goodhue} {and}
  \bibinfo{person}{Detmar~W Straub}.} \bibinfo{year}{1991}\natexlab{}.
\newblock \showarticletitle{Security concerns of system users: A study of
  perceptions of the adequacy of security}.
\newblock \bibinfo{journal}{\emph{Information \& Management}}
  \bibinfo{volume}{20}, \bibinfo{number}{1} (\bibinfo{year}{1991}),
  \bibinfo{pages}{13--27}.
\newblock


\bibitem[\protect\citeauthoryear{Hayes and Capretz}{Hayes and Capretz}{2015}]%
        {hayes2015contextual}
\bibfield{author}{\bibinfo{person}{Michael~A Hayes} {and}
  \bibinfo{person}{Miriam~AM Capretz}.} \bibinfo{year}{2015}\natexlab{}.
\newblock \showarticletitle{Contextual anomaly detection framework for big
  sensor data}.
\newblock \bibinfo{journal}{\emph{Journal of Big Data}} \bibinfo{volume}{2},
  \bibinfo{number}{1} (\bibinfo{year}{2015}), \bibinfo{pages}{2}.
\newblock


\bibitem[\protect\citeauthoryear{Hochreiter and Schmidhuber}{Hochreiter and
  Schmidhuber}{1997}]%
        {HochSchm97}
\bibfield{author}{\bibinfo{person}{Sepp Hochreiter} {and}
  \bibinfo{person}{Jürgen Schmidhuber}.} \bibinfo{year}{1997}\natexlab{}.
\newblock \showarticletitle{Long short-term memory}.
\newblock \bibinfo{journal}{\emph{Neural Computation}} \bibinfo{volume}{9},
  \bibinfo{number}{8} (\bibinfo{year}{1997}), \bibinfo{pages}{1735--1780}.
\newblock


\bibitem[\protect\citeauthoryear{Id{\'e}, Papadimitriou, and Vlachos}{Id{\'e}
  et~al\mbox{.}}{2007}]%
        {ide2007computing}
\bibfield{author}{\bibinfo{person}{Tsuyoshi Id{\'e}}, \bibinfo{person}{Spiros
  Papadimitriou}, {and} \bibinfo{person}{Michail Vlachos}.}
  \bibinfo{year}{2007}\natexlab{}.
\newblock \showarticletitle{Computing correlation anomaly scores using
  stochastic nearest neighbors}. In \bibinfo{booktitle}{\emph{Proc. IEEE
  International Conference on Data Mining}}. \bibinfo{pages}{523--528}.
\newblock


\bibitem[\protect\citeauthoryear{Kazemi, Vesilo, and Dutkiewicz}{Kazemi
  et~al\mbox{.}}{2011}]%
        {kazemi2011novel}
\bibfield{author}{\bibinfo{person}{Ramtin Kazemi}, \bibinfo{person}{Rein
  Vesilo}, {and} \bibinfo{person}{Eryk Dutkiewicz}.}
  \bibinfo{year}{2011}\natexlab{}.
\newblock \showarticletitle{A novel genetic-fuzzy power controller with
  feedback for interference mitigation in wireless body area networks}. In
  \bibinfo{booktitle}{\emph{Proc. IEEE Vehicular Technology Conference}}.
  \bibinfo{pages}{1--5}.
\newblock


\bibitem[\protect\citeauthoryear{Khamesi, Silvestri, Baker, and {De
  Paola}}{Khamesi et~al\mbox{.}}{2020}]%
        {Khamesi:20}
\bibfield{author}{\bibinfo{person}{A.R. Khamesi}, \bibinfo{person}{S.
  Silvestri}, \bibinfo{person}{D.A. Baker}, {and} \bibinfo{person}{A. {De
  Paola}}.} \bibinfo{year}{2020}\natexlab{}.
\newblock \showarticletitle{Perceived-Value-driven Optimization of Energy
  Consumption in Smart Homes}.
\newblock \bibinfo{journal}{\emph{ACM Transactions on Internet of Things}}
  \bibinfo{volume}{1} (\bibinfo{year}{2020}).
\newblock


\bibitem[\protect\citeauthoryear{Kim, Kang, Broman, and Lee}{Kim
  et~al\mbox{.}}{2020}]%
        {HKim:20}
\bibfield{author}{\bibinfo{person}{Hokeun Kim}, \bibinfo{person}{Eunsuk Kang},
  \bibinfo{person}{David Broman}, {and} \bibinfo{person}{Edward~A. Lee}.}
  \bibinfo{year}{2020}\natexlab{}.
\newblock \showarticletitle{Resilient Authentication and Authorization for the
  Internet of Things ({IoT}) Using Edge Computing}.
\newblock \bibinfo{journal}{\emph{ACM Transactions on Internet of Things}}
  \bibinfo{volume}{1} (\bibinfo{year}{2020}).
\newblock


\bibitem[\protect\citeauthoryear{Konecn\'{y}, McMahan, Yu, Richt{\'{a}}rik,
  Suresh, Bacon, and Richt\'{a}rik}{Konecn\'{y} et~al\mbox{.}}{2016}]%
        {KonecnyMYRSB16}
\bibfield{author}{\bibinfo{person}{J. Konecn\'{y}}, \bibinfo{person}{H.B.
  McMahan}, \bibinfo{person}{F.X. Yu}, \bibinfo{person}{P. Richt{\'{a}}rik},
  \bibinfo{person}{A.T. Suresh}, \bibinfo{person}{D. Bacon}, {and}
  \bibinfo{person}{P. Richt\'{a}rik}.} \bibinfo{year}{2016}\natexlab{}.
\newblock \showarticletitle{Federated learning: strategies for improving
  communication efficiency}. In \bibinfo{booktitle}{\emph{NIPS Workshop on
  Private Multi-Party Machine Learning}}. \bibinfo{pages}{322--334}.
\newblock


\bibitem[\protect\citeauthoryear{Li, Cheng, Liu, Wang, and Chen}{Li
  et~al\mbox{.}}{2019}]%
        {li2019abnormal}
\bibfield{author}{\bibinfo{person}{Suyi Li}, \bibinfo{person}{Yong Cheng},
  \bibinfo{person}{Yang Liu}, \bibinfo{person}{Wei Wang}, {and}
  \bibinfo{person}{Tianjian Chen}.} \bibinfo{year}{2019}\natexlab{}.
\newblock \showarticletitle{Abnormal client behavior detection in federated
  learning}. In \bibinfo{booktitle}{\emph{NeurIPS Workshop on Federated
  Learning for Data Privacy and Confidentiality}}. \bibinfo{pages}{740--750}.
\newblock


\bibitem[\protect\citeauthoryear{Li, Song, and Zhou}{Li et~al\mbox{.}}{2018}]%
        {li2018leak}
\bibfield{author}{\bibinfo{person}{Suzhen Li}, \bibinfo{person}{Yanjue Song},
  {and} \bibinfo{person}{Gongqi Zhou}.} \bibinfo{year}{2018}\natexlab{}.
\newblock \showarticletitle{Leak detection of water distribution pipeline
  subject to failure of socket joint based on acoustic emission and pattern
  recognition}.
\newblock \bibinfo{journal}{\emph{Measurement}}  \bibinfo{volume}{115}
  (\bibinfo{year}{2018}), \bibinfo{pages}{39--44}.
\newblock


\bibitem[\protect\citeauthoryear{Li, Sahu, Talwalkar, and Smith}{Li
  et~al\mbox{.}}{2020}]%
        {TianLi:19}
\bibfield{author}{\bibinfo{person}{Tian Li}, \bibinfo{person}{Anit~Kumar Sahu},
  \bibinfo{person}{Ameet Talwalkar}, {and} \bibinfo{person}{Virginia Smith}.}
  \bibinfo{year}{2020}\natexlab{}.
\newblock \showarticletitle{Federated Learning: Challenges, Methods, and Future
  Directions}.
\newblock \bibinfo{journal}{\emph{IEEE Signal Processing Magazine}}
  \bibinfo{volume}{37}, \bibinfo{number}{3} (\bibinfo{year}{2020}),
  \bibinfo{pages}{5--60}.
\newblock


\bibitem[\protect\citeauthoryear{Liang, Liu, Ziyin, Salakhutdinov, and
  Morency}{Liang et~al\mbox{.}}{2018}]%
        {PLiang:20}
\bibfield{author}{\bibinfo{person}{P.P. Liang}, \bibinfo{person}{T. Liu},
  \bibinfo{person}{L. Ziyin}, \bibinfo{person}{R. Salakhutdinov}, {and}
  \bibinfo{person}{L.-P. Morency}.} \bibinfo{year}{2018}\natexlab{}.
\newblock \showarticletitle{Think Locally, Act Globally: Federated Learning
  with Local and Global Representations}. In \bibinfo{booktitle}{\emph{NeurIPS
  Workshop on Federated Learning for Data Privacy and Confidentiality}}.
\newblock


\bibitem[\protect\citeauthoryear{Lim, Luong, Hoang, Jiao, Liang, Yang, Niyato,
  and Miao}{Lim et~al\mbox{.}}{2020}]%
        {Niyato:19}
\bibfield{author}{\bibinfo{person}{W.Y.B. Lim}, \bibinfo{person}{N.C. Luong},
  \bibinfo{person}{D.T. Hoang}, \bibinfo{person}{Y. Jiao},
  \bibinfo{person}{Y.-C. Liang}, \bibinfo{person}{Q. Yang}, \bibinfo{person}{D.
  Niyato}, {and} \bibinfo{person}{C. Miao}.} \bibinfo{year}{2020}\natexlab{}.
\newblock \showarticletitle{Federated Learning in Mobile Edge Networks: A
  Comprehensive Survey}.
\newblock \bibinfo{journal}{\emph{IEEE Communications Surveys and Tutorials}}
  \bibinfo{volume}{22}, \bibinfo{number}{3} (\bibinfo{year}{2020}),
  \bibinfo{pages}{2031--2063}.
\newblock


\bibitem[\protect\citeauthoryear{Ma, Liu, Wang, Zhang, Lei, Zeng, Wang, and
  Cheng}{Ma et~al\mbox{.}}{2016}]%
        {ma2016multi}
\bibfield{author}{\bibinfo{person}{Li Ma}, \bibinfo{person}{Nian Liu},
  \bibinfo{person}{Lingfeng Wang}, \bibinfo{person}{Jianhua Zhang},
  \bibinfo{person}{Jinyong Lei}, \bibinfo{person}{Zheng Zeng},
  \bibinfo{person}{Cheng Wang}, {and} \bibinfo{person}{Minyang Cheng}.}
  \bibinfo{year}{2016}\natexlab{}.
\newblock \showarticletitle{Multi-party energy management for smart building
  cluster with {PV} systems using automatic demand response}.
\newblock \bibinfo{journal}{\emph{Energy and {B}uildings}}
  \bibinfo{volume}{121} (\bibinfo{year}{2016}), \bibinfo{pages}{11--21}.
\newblock


\bibitem[\protect\citeauthoryear{McMahan, Ramage, Talwar, and Zhang}{McMahan
  et~al\mbox{.}}{2018}]%
        {brendan2017learning}
\bibfield{author}{\bibinfo{person}{H.B. McMahan}, \bibinfo{person}{D. Ramage},
  \bibinfo{person}{K. Talwar}, {and} \bibinfo{person}{L. Zhang}.}
  \bibinfo{year}{2018}\natexlab{}.
\newblock \showarticletitle{Learning differentially private recurrent language
  models}. In \bibinfo{booktitle}{\emph{International Conference on Learning
  Representations}}.
\newblock


\bibitem[\protect\citeauthoryear{McMahan, Moore, Ramage, Hampson, and
  y~Arcas}{McMahan et~al\mbox{.}}{2017}]%
        {mcmahan2016communication}
\bibfield{author}{\bibinfo{person}{H~Brendan McMahan}, \bibinfo{person}{Eider
  Moore}, \bibinfo{person}{Daniel Ramage}, \bibinfo{person}{Seth Hampson},
  {and} \bibinfo{person}{Blaise~Ag\"{u}era y Arcas}.}
  \bibinfo{year}{2017}\natexlab{}.
\newblock \showarticletitle{Communication-efficient learning of deep networks
  from decentralized data}. In \bibinfo{booktitle}{\emph{International
  Conference on Artificial Intelligence and Statistics}}.
\newblock


\bibitem[\protect\citeauthoryear{Mitchell}{Mitchell}{2007}]%
        {mitchell2007multi}
\bibfield{author}{\bibinfo{person}{Harvey~B Mitchell}.}
  \bibinfo{year}{2007}\natexlab{}.
\newblock \bibinfo{booktitle}{\emph{Multi-Sensor Data Fusion: An
  Introduction}}.
\newblock \bibinfo{publisher}{Springer Science \& Business Media}.
\newblock


\bibitem[\protect\citeauthoryear{Mohri, Sivek, and Suresh}{Mohri
  et~al\mbox{.}}{2019}]%
        {Mehryar:19}
\bibfield{author}{\bibinfo{person}{Mehryar Mohri}, \bibinfo{person}{Gary
  Sivek}, {and} \bibinfo{person}{Ananda~Theertha Suresh}.}
  \bibinfo{year}{2019}\natexlab{}.
\newblock \showarticletitle{Agnostic Federated Learning}. In
  \bibinfo{booktitle}{\emph{Proc. International Conference on Machine
  Learning}}.
\newblock


\bibitem[\protect\citeauthoryear{Mosenia and Jha}{Mosenia and Jha}{2017}]%
        {Mosenia:17}
\bibfield{author}{\bibinfo{person}{A. Mosenia} {and} \bibinfo{person}{N.K.
  Jha}.} \bibinfo{year}{2017}\natexlab{}.
\newblock \showarticletitle{A Comprehensive Study of Security of
  Internet-of-Things}.
\newblock \bibinfo{journal}{\emph{IEEE Transactions on Emerging Topics in
  Computing}}  \bibinfo{volume}{5} (\bibinfo{year}{2017}),
  \bibinfo{pages}{586--602}.
\newblock


\bibitem[\protect\citeauthoryear{Nguyen, Marchal, Miettinen, Fereidooni,
  Asokan, and Sadeghi}{Nguyen et~al\mbox{.}}{2019}]%
        {nguyen2019diot}
\bibfield{author}{\bibinfo{person}{Thien~Duc Nguyen}, \bibinfo{person}{Samuel
  Marchal}, \bibinfo{person}{Markus Miettinen}, \bibinfo{person}{Hossein
  Fereidooni}, \bibinfo{person}{N Asokan}, {and} \bibinfo{person}{Ahmad-Reza
  Sadeghi}.} \bibinfo{year}{2019}\natexlab{}.
\newblock \showarticletitle{{D{\"I}oT}: A federated self-learning anomaly
  detection system for {IoT}}. In \bibinfo{booktitle}{\emph{Proc. IEEE
  International Conference on Distributed Computing Systems}}.
  \bibinfo{pages}{756--767}.
\newblock


\bibitem[\protect\citeauthoryear{Pajouh, Javidan, Khayami, Dehghantanha, and
  Choo}{Pajouh et~al\mbox{.}}{2019}]%
        {Pajouh:19}
\bibfield{author}{\bibinfo{person}{H.H. Pajouh}, \bibinfo{person}{R. Javidan},
  \bibinfo{person}{R. Khayami}, \bibinfo{person}{A. Dehghantanha}, {and}
  \bibinfo{person}{K.-K.R. Choo}.} \bibinfo{year}{2019}\natexlab{}.
\newblock \showarticletitle{A Two-Layer Dimension Reduction and Two-Tier
  Classification Model for Anomaly-Based Intrusion Detection in IoT Backbone
  Networks}.
\newblock \bibinfo{journal}{\emph{IEEE Transactions on Emerging Topics in
  Computing}}  \bibinfo{volume}{7} (\bibinfo{year}{2019}),
  \bibinfo{pages}{314--323}.
\newblock


\bibitem[\protect\citeauthoryear{Perera, Liu, and Jayawardena}{Perera
  et~al\mbox{.}}{2015}]%
        {Perera:15}
\bibfield{author}{\bibinfo{person}{C. Perera}, \bibinfo{person}{C.H. Liu},
  {and} \bibinfo{person}{S. Jayawardena}.} \bibinfo{year}{2015}\natexlab{}.
\newblock \showarticletitle{The Emerging Internet of Things Marketplace From an
  Industrial Perspective: A Survey}.
\newblock \bibinfo{journal}{\emph{IEEE Transactions on Emerging Topics in
  Computing}}  \bibinfo{volume}{3} (\bibinfo{year}{2015}),
  \bibinfo{pages}{585--598}.
\newblock


\bibitem[\protect\citeauthoryear{Reddi, Charles, Zaheer, Garrett, Rush,
  Konecny, Kumar, and McMahan}{Reddi et~al\mbox{.}}{2021}]%
        {Sashank:2021}
\bibfield{author}{\bibinfo{person}{Sashank Reddi}, \bibinfo{person}{Zachary
  Charles}, \bibinfo{person}{Manzil Zaheer}, \bibinfo{person}{Zachary Garrett},
  \bibinfo{person}{Keith Rush}, \bibinfo{person}{Jakub Konecny},
  \bibinfo{person}{Sanjiv Kumar}, {and} \bibinfo{person}{H.~Brendan McMahan}.}
  \bibinfo{year}{2021}\natexlab{}.
\newblock \showarticletitle{Adaptive Federated Optimization}. In
  \bibinfo{booktitle}{\emph{International Conference on Learning
  Representations}}.
\newblock


\bibitem[\protect\citeauthoryear{Ryffel, Trask, Dahl, Wagner, Mancuso,
  Rueckert, and Passerat-Palmbach}{Ryffel et~al\mbox{.}}{2018}]%
        {PySyft:18}
\bibfield{author}{\bibinfo{person}{Theo Ryffel}, \bibinfo{person}{Andrew
  Trask}, \bibinfo{person}{Morten Dahl}, \bibinfo{person}{Bobby Wagner},
  \bibinfo{person}{Jason Mancuso}, \bibinfo{person}{Daniel Rueckert}, {and}
  \bibinfo{person}{Jonathan Passerat-Palmbach}.}
  \bibinfo{year}{2018}\natexlab{}.
\newblock \showarticletitle{A generic framework for privacy preserving deep
  learning}. In \bibinfo{booktitle}{\emph{Proc. NeurIPS Workshop in Privacy
  Preserving Machine Learning}}.
\newblock


\bibitem[\protect\citeauthoryear{Violatto, Pandharipande, Li, and
  Schenato}{Violatto et~al\mbox{.}}{2019}]%
        {violatto2019classification}
\bibfield{author}{\bibinfo{person}{Giulia Violatto}, \bibinfo{person}{Ashish
  Pandharipande}, \bibinfo{person}{Shuai Li}, {and} \bibinfo{person}{Luca
  Schenato}.} \bibinfo{year}{2019}\natexlab{}.
\newblock \showarticletitle{Classification of Occupancy Sensor Anomalies in
  Connected Indoor Lighting Systems}.
\newblock \bibinfo{journal}{\emph{{IEEE} {I}nternet of {T}hings {J}ournal}}
  \bibinfo{volume}{6}, \bibinfo{number}{4} (\bibinfo{year}{2019}),
  \bibinfo{pages}{7175--7182}.
\newblock


\bibitem[\protect\citeauthoryear{Wang, Yurochkin, Sun, Papailiopoulos, and
  Khazaeni}{Wang et~al\mbox{.}}{2020}]%
        {Hongyi:20}
\bibfield{author}{\bibinfo{person}{Hongyi Wang}, \bibinfo{person}{Mikhail
  Yurochkin}, \bibinfo{person}{Yuekai Sun}, \bibinfo{person}{Dimitris
  Papailiopoulos}, {and} \bibinfo{person}{Yasaman Khazaeni}.}
  \bibinfo{year}{2020}\natexlab{}.
\newblock \showarticletitle{Federated Learning with Matched Averaging}. In
  \bibinfo{booktitle}{\emph{International Conference on Learning
  Representations}}.
\newblock


\bibitem[\protect\citeauthoryear{Yao, Huang, and Sun}{Yao
  et~al\mbox{.}}{2018}]%
        {yao2018two}
\bibfield{author}{\bibinfo{person}{Xin Yao}, \bibinfo{person}{Chaofeng Huang},
  {and} \bibinfo{person}{Lifeng Sun}.} \bibinfo{year}{2018}\natexlab{}.
\newblock \showarticletitle{Two-stream federated learning: Reduce the
  communication costs}. In \bibinfo{booktitle}{\emph{Proc. IEEE Visual
  Communications and Image Processing}}. \bibinfo{pages}{1--4}.
\newblock


\bibitem[\protect\citeauthoryear{Yu, Xie, Jiang, Zou, and Wang}{Yu
  et~al\mbox{.}}{2017}]%
        {yu2017distributed}
\bibfield{author}{\bibinfo{person}{Liang Yu}, \bibinfo{person}{Di Xie},
  \bibinfo{person}{Tao Jiang}, \bibinfo{person}{Yulong Zou}, {and}
  \bibinfo{person}{Kun Wang}.} \bibinfo{year}{2017}\natexlab{}.
\newblock \showarticletitle{Distributed real-time {HVAC} control for
  cost-efficient commercial buildings under smart grid environment}.
\newblock \bibinfo{journal}{\emph{IEEE Internet of Things Journal}}
  \bibinfo{volume}{5}, \bibinfo{number}{1} (\bibinfo{year}{2017}),
  \bibinfo{pages}{44--55}.
\newblock


\bibitem[\protect\citeauthoryear{Yurochkin, Agarwal, Ghosh, Greenewald, Hoang,
  and Khazaeni}{Yurochkin et~al\mbox{.}}{2019}]%
        {Yurochkin:19}
\bibfield{author}{\bibinfo{person}{Mikhail Yurochkin}, \bibinfo{person}{Mayank
  Agarwal}, \bibinfo{person}{Soumya Ghosh}, \bibinfo{person}{Kristjan
  Greenewald}, \bibinfo{person}{Nghia Hoang}, {and} \bibinfo{person}{Yasaman
  Khazaeni}.} \bibinfo{year}{2019}\natexlab{}.
\newblock \showarticletitle{Bayesian Nonparametric Federated Learning of Neural
  Networks}. In \bibinfo{booktitle}{\emph{Proc. International Cconference on
  Machine Learning}}.
\newblock


\bibitem[\protect\citeauthoryear{Zhang, Guo, Liu, Liu, Zhou, Li, Lu, and
  Yang}{Zhang et~al\mbox{.}}{2018}]%
        {zhang2018lstm}
\bibfield{author}{\bibinfo{person}{Weishan Zhang}, \bibinfo{person}{Wuwu Guo},
  \bibinfo{person}{Xin Liu}, \bibinfo{person}{Yan Liu}, \bibinfo{person}{Jiehan
  Zhou}, \bibinfo{person}{Bo Li}, \bibinfo{person}{Qinghua Lu}, {and}
  \bibinfo{person}{Su Yang}.} \bibinfo{year}{2018}\natexlab{}.
\newblock \showarticletitle{{LSTM}-based analysis of industrial {I}o{T}
  equipment}.
\newblock \bibinfo{journal}{\emph{IEEE Access}}  \bibinfo{volume}{6}
  (\bibinfo{year}{2018}), \bibinfo{pages}{23551--23560}.
\newblock


\bibitem[\protect\citeauthoryear{Zhao, Chen, and Wu}{Zhao
  et~al\mbox{.}}{2019}]%
        {YZhao:19}
\bibfield{author}{\bibinfo{person}{Y. Zhao}, \bibinfo{person}{J. Chen}, {and}
  \bibinfo{person}{D. Wu}.} \bibinfo{year}{2019}\natexlab{}.
\newblock \showarticletitle{Multi-Task Network Anomaly Detection using
  Federated Learning}. In \bibinfo{booktitle}{\emph{Proc. International
  Symposium on Information and Communication Technology}}.
\newblock


\bibitem[\protect\citeauthoryear{Zheng, Zhang, and Yu}{Zheng
  et~al\mbox{.}}{2015}]%
        {zheng2015detecting}
\bibfield{author}{\bibinfo{person}{Yu Zheng}, \bibinfo{person}{Huichu Zhang},
  {and} \bibinfo{person}{Yong Yu}.} \bibinfo{year}{2015}\natexlab{}.
\newblock \showarticletitle{Detecting collective anomalies from multiple
  spatio-temporal datasets across different domains}. In
  \bibinfo{booktitle}{\emph{Proc. SIGSPATIAL International Conference on
  Advances in Geographic Information Systems}}. \bibinfo{pages}{1--10}.
\newblock


\bibitem[\protect\citeauthoryear{Zhu, Chen, Peng, and Zhang}{Zhu
  et~al\mbox{.}}{2019a}]%
        {zhu2019mobile}
\bibfield{author}{\bibinfo{person}{Konglin Zhu}, \bibinfo{person}{Zhicheng
  Chen}, \bibinfo{person}{Yuyang Peng}, {and} \bibinfo{person}{Lin Zhang}.}
  \bibinfo{year}{2019}\natexlab{a}.
\newblock \showarticletitle{Mobile edge assisted literal multi-dimensional
  anomaly detection of in-vehicle network using {LSTM}}.
\newblock \bibinfo{journal}{\emph{IEEE {T}ransactions on {V}ehicular
  {T}echnology}} \bibinfo{volume}{68}, \bibinfo{number}{5}
  (\bibinfo{year}{2019}), \bibinfo{pages}{4275--4284}.
\newblock


\bibitem[\protect\citeauthoryear{Zhu, Liu, and Han}{Zhu et~al\mbox{.}}{2019b}]%
        {zhu2019dlg}
\bibfield{author}{\bibinfo{person}{Ligeng Zhu}, \bibinfo{person}{Zhijian Liu},
  {and} \bibinfo{person}{Song Han}.} \bibinfo{year}{2019}\natexlab{b}.
\newblock \showarticletitle{Deep Leakage from Gradients}. In
  \bibinfo{booktitle}{\emph{Advances in Neural Information Processing
  Systems}}.
\newblock


\end{thebibliography}


\end{document}